\documentclass[12pt]{scaleai-paper}
\usepackage[square,sort&compress,numbers]{natbib}
\usepackage{mathpazo}

\usepackage[scaled=0.92]{helvet} 
\usepackage{fontawesome5}
\usepackage{subcaption}
\usepackage{hyperref}
\usepackage{tcolorbox}
\usepackage{graphicx}
\usepackage{booktabs,makecell}
\usepackage{fontawesome5}

\usepackage{hyperref}
\hypersetup{
    colorlinks=true,
	linkcolor=blue,
	filecolor=magenta,      
	urlcolor=blue,
	citecolor=black,
	pdfinfo={
        Title={Evaluating Frontier Models for Legal and Finance},
        Subject={finance benchmark, legal benchmark, llms},
        Keywords={legal, finance, professional domains, scale ai, rubrics},
    }
}
\usepackage{longtable}
\usepackage{tabularx}
\usepackage[utf8]{inputenc} 
\usepackage[T1]{fontenc}    
\usepackage{hyperref}       
\usepackage{url}            
\usepackage{booktabs}       
\usepackage{amsfonts}       
\usepackage{nicefrac}       
\usepackage{microtype}      
\usepackage{booktabs}
\usepackage{multirow}
\usepackage{amsmath}
\usepackage[ruled,vlined]{algorithm2e}
\usepackage[capitalise]{cleveref}
\usepackage{paralist}
\usepackage{amsthm}
\usepackage{xspace}
\usepackage{graphicx}
\usepackage[table]{xcolor}
\usepackage{array}

\usepackage{enumitem}
\setlist[itemize]{leftmargin=*}
\setlist[enumerate]{leftmargin=*}

\usepackage{graphicx}
\usepackage{tcolorbox}

\usepackage{amsmath}
\usepackage{soul}
\usepackage{cleveref}
\usepackage{listings}
\usepackage{tikz}
\usepackage{eso-pic}
\usepackage{wrapfig}
\usepackage{pifont}

\usepackage{amsmath,amsfonts,bm}

\def\eqref#1{equation~\ref{#1}}

\def\1{\bm{1}}

\DeclareMathAlphabet{\mathsfit}{\encodingdefault}{\sfdefault}{m}{sl}
\SetMathAlphabet{\mathsfit}{bold}{\encodingdefault}{\sfdefault}{bx}{n}

\newcommand{\cinterval}[1]{\ensuremath{\left[#1\right]}}

\definecolor{ao}{rgb}{0.0, 0.5, 0.0}
\definecolor{red}{rgb}{1.0, 0.0, 0.0}

\let\svthefootnote\thefootnote
\newcommand\freefootnote[1]{%
  \let\thefootnote\relax%
  \footnotetext{#1}%
  \let\thefootnote\svthefootnote%
}

\makeatletter
\renewcommand\AB@affilsepx{, \protect\Affilfont}
\makeatother

\newcommand{\benchmarkname}{PRBench\xspace}

\title{\benchmarkname: Large-Scale Expert Rubrics for Evaluating High-Stakes \underline{P}rofessional \underline{R}easoning}

\author{Afra Feyza Akyürek, Advait Gosai, Chen Bo Calvin Zhang, Vipul Gupta, Jaehwan Jeong, Anisha Gunjal, Tahseen Rabbani, Maria Mazzone, David Randolph, Mohammad Mahmoudi Meymand, Gurshaan Chattha, Paula Rodriguez, Diego Mares, Pavit Singh, Michael Liu, Subodh Chawla, Pete Cline, Lucy Ogaz, Ernesto Hernandez, Zihao Wang, Pavi Bhatter, Marcos Ayestaran, Bing Liu and Yunzhong He}

\affil{Scale AI}

\newcommand{\authoremail}{%
  \vspace{-1.5em}
    \faEnvelope\  \texttt{yunzhong.he@scale.com} \quad 
    \faGlobe\  \url{https://scale.com/research/prbench}
}

\let\oldcite\cite
\let\cite\citet
\let\citep\oldcite

\crefname{proposition}{Proposition}{Propositions}

\crefname{figure}{Figure}{Figures}
\Crefname{figure}{Figure}{Figures}

\crefname{table}{Table}{Tables}
\Crefname{table}{Table}{Tables}

\crefname{section}{Section}{Sections}
\Crefname{section}{Section}{Sections}

\crefname{equation}{Equation}{Equations}
\Crefname{equation}{Equation}{Equations}
\crefname{appendix}{Appendix}{Appendices}

\begin{document}
\maketitle
\authoremail

\begin{abstract}
Frontier model progress is often measured by academic benchmarks, which offer a limited view of performance in real-world professional contexts. Existing evaluations often fail to assess open-ended, economically consequential tasks in high-stakes domains like Legal and Finance, where practical returns are paramount. To address this, we introduce \textbf{Professional Reasoning Bench} (\textbf{PRBench}), a realistic, open-ended, and difficult benchmark of real-world problems in Finance and Law. We open-source its 1,100 expert-authored tasks and 19,356 expert-curated criteria, making it, to our knowledge, the largest public, rubric-based benchmark for both legal and finance domains. We recruit 182 qualified professionals, holding JDs, CFAs, or 6+ years of experience, who contributed tasks inspired by their actual workflows. This process yields significant diversity, with tasks spanning 114 countries and 47 US jurisdictions. Our expert-curated rubrics are validated through a rigorous quality pipeline, including independent expert validation. Subsequent evaluation of 20 leading models reveals substantial room for improvement, with top scores of only 0.39 (Finance) and 0.37 (Legal) on our Hard subsets. We further catalog associated economic impacts of the prompts and analyze performance using human-annotated rubric categories. Our analysis shows that models with similar overall scores can diverge significantly on specific capabilities. Common failure modes include inaccurate judgments, a lack of process transparency and incomplete reasoning, highlighting critical gaps in their reliability for professional adoption.
\begin{figure}[!h]
    \centering
    \includegraphics[width=0.95\textwidth]{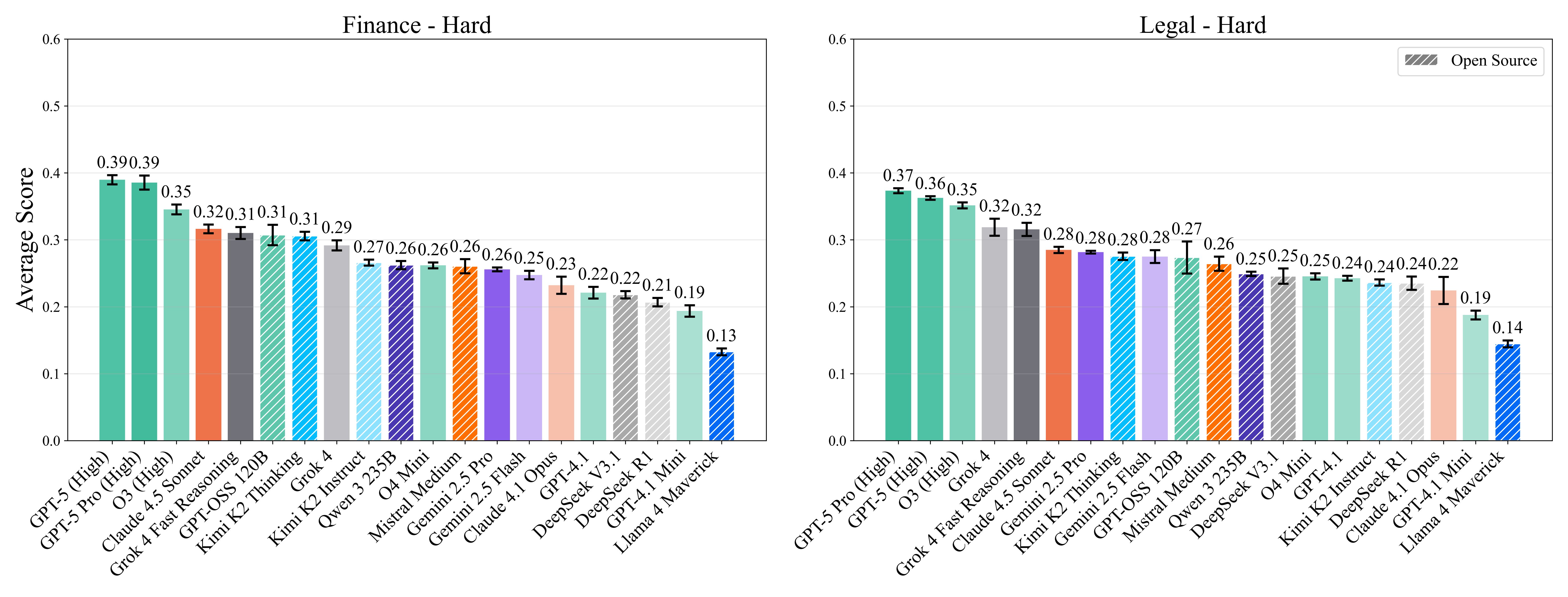}
    \caption{Results on Legal and Finance Hard subsets of \benchmarkname.}
    \label{fig:hard_subsets}
\end{figure}
\end{abstract}

\section{Introduction} \label{sec:introduction}

Frontier chat model progress has traditionally been measured using benchmarks focused on reasoning tasks with verifiable answers, primarily across mathematics, science, and coding domains. Prominent examples include GPQA~\citep{rein2024gpqa}, ARC-AGI~\citep{chollet2025arc}, MMLU~\citep{hendrycks2021mmlu}, AIME and Humanity's Last Exam~\citep{kim2025hle}, which collectively assess academic and scientific reasoning. While these evaluations have become the \textit{de facto} metrics for tracking advances in general reasoning ability, they offer a limited view of how such models perform in real-world professional contexts. The pace at which frontier models improve on these academic-style benchmarks contrasts sharply with the comparatively modest returns in practical or business applications, particularly for expert use cases that demand domain knowledge, judgment, and contextual nuance.

While math and science reasoning tasks effectively measure an LLM's competence in STEM reasoning, they offer limited visibility into performance in real-world professional domains. This creates a gap between what current evaluations measure and the capabilities needed to produce economic and professional impact. Recent usage data reinforce this gap. Although work-related queries are increasing steadily~\citep{chatterji2025people,appel2025anthropic}, they still lag behind non-work interactions, suggesting both a growing desire to integrate LLMs into reasoning and decision-making workflows and a lingering lack of trust or perceived utility.

Another important dimension that current benchmarks overlook is the evaluation of open-ended tasks. Most existing evaluations for such tasks rely on preference-ranking benchmarks and public arenas such as Chatbot Arena, AlpacaEval~\citep{alpaca_eval}, and more recently Showdown~\citep{scale_showdown_2025} and Arena Expert~\citep{LMArena2025ArenaExpert}. While these provide a useful aggregate signal of user preference, they remain coarse and difficult to interpret. The resulting scores are often noisy, subjective, and lack expert grounding, making it challenging to derive actionable insights about model capabilities. These findings underscore the limitations of current evaluation paradigms for capturing domain-specific reasoning quality in open-ended contexts.

Usage analyses from Anthropic's \texttt{claude.ai} identify Legal and Business \& Financial Operations among the most common professional categories of interaction~\citep{appel2025anthropic}. Similarly, OpenAI reports that Legal and Business/Management tasks rank among the top work activities on \texttt{chatgpt.com}. These domains are not only among the most frequent professional use cases but also among the most high-stakes, where reasoning quality, factual precision, and interpretability directly affect real-world outcomes, including financial outcomes and user trust. Yet, they remain largely underexplored in systematic evaluation efforts.

To address this gap, we introduce \textbf{Professional Reasoning Bench} (\textbf{\benchmarkname}), a suite of 1,100 expert-authored questions designed to evaluate reasoning-heavy, real-world problems across 114 countries for \textbf{Legal} and \textbf{Finance} domains. Questions are derived from experts' actual experiences using chat-based assistants, as well as the types of inquiries they commonly receive from clients. Each question is accompanied by an expert-curated and verified rubric containing 10--30 descriptive criteria with importance weights, enabling automated and interpretable evaluation. Following the methodology of HealthBench~\citep{arora2025healthbench}, we additionally identify a Hard subset of 250 and 300 questions for the legal and finance domains, respectively, representing the most challenging cases. Current best scores remain at only 0.39 and 0.37 for Finance and Legal subsets, respectively, highlighting significant headroom for improvement in these domains. 

\benchmarkname provides substantial improvements over existing benchmarks in professional domains~\citep{guha2023legalbench,biglaw,ValsAI2025CorpFinV2} by being realistic, open-ended, and difficult, where existing benchmarks are near-saturated, focus on narrowly defined tasks, or rely on non-interpretable evaluation methods. Furthermore, existing rubric-based evaluations for professional tasks are often limited by being private or small in scale, which restricts accessibility and comprehensive coverage \citep{apex, profbench, biglaw}. We address this by open-sourcing \benchmarkname, which, with 1,100 tasks and 19,356 expert-curated criteria, is the largest public, rubric-based benchmark for both legal and finance domains to our knowledge.

Our analysis reveals that while LLMs tend to perform better on instruction following and practical utility dimensions compared to other aspects, they continue to struggle with process transparency, auditability, correctness, and domain-specific diligence. Models frequently make inaccurate legal or financial judgments or reach correct conclusions through incomplete or opaque reasoning processes, reducing their practical reliability and slowing professional adoption. Furthermore, we qualitatively analyze both prompts and rubrics to identify systematic areas for improvement and to recommend concrete paths for model development.
Our contributions are as follows:
\begin{itemize}
    \item We open-source\footnote{ \href{https://huggingface.co/datasets/ScaleAI/PRBench}{https://huggingface.co/datasets/ScaleAI/PRBench}} a total of 1,100 realistic, challenging tasks for evaluating frontier LLM-based chat assistants covering 25 topics in Finance and Legal domains. Prompts in \benchmarkname cover 114 countries and dependencies globally and 47 jurisdictions within the US.
    \item Each task is evaluated with an expert-curated rubric comprising detailed and diverse criteria. Across two benchmarks, we are releasing a total of 19,356 criteria over 11 rubric categories, making this the largest public, rubric-based benchmark for both legal and finance domains to our knowledge.
    \item We assess the reliability of \benchmarkname through a rubric validation study with an independent set of domain experts, complementing our internal quality checks to ensure the robustness of rubric design.
    \item We evaluate\footnote{\href{https://github.com/scaleapi/PRBench}{https://github.com/scaleapi/PRBench}} the performance of open-source and proprietary chat models in our benchmarks, showing that substantial room for improvement remains.
    \item We provide an analysis of the types of prompts and rubrics included in this dataset: we annotate the economic implications associated with different prompt types and the decision types which can be used to analyze where LLMs tend to perform well or poorly.
\end{itemize}

\section{Overview of \benchmarkname} \label{sec:overview}
\begin{wraptable}{r}{0.45\textwidth}
\vspace{-2em}
\centering
\small
\caption{Dataset Statistics for \benchmarkname.}
\label{tab:dataset_stats}
\begin{tabular}{llcc}
\toprule
 & & \textbf{Finance} & \textbf{Law} \\
\midrule
 & Total Samples & 600 & 500 \\
 & Hard Subset & 300 & 250 \\
\midrule
User Expertise & Expert & 74\% & 53\% \\
 & Non-Expert & 26\% & 47\% \\
\midrule
\# of Rubrics & Min & 10 & 10 \\
 & 25\% & 13 & 15 \\
 & Median & 16 & 17 \\
 & 75\% & 20 & 22 \\
 & Max & 30 & 30 \\
 & \textit{Total} & 10264 & 9092 \\
\midrule
Turns & Min & 1 & 1 \\
 & 25\% & 1 & 1 \\
 & Median & 1 & 1 \\
 & 75\% & 2 & 2 \\
 & Max & 10 & 10 \\
\bottomrule
\end{tabular}
\end{wraptable}

Overall statistics for the two benchmarks are given in \cref{tab:dataset_stats}. All conversations in the benchmark are created by a set of 182 experts across two domains who have passed through resume checks and internal qualification assessments. Approximately 30\% of all conversations in the dataset are multi-turn. All \textit{user} turns are written in English by human experts, and \textit{assistant} turns are sampled from one of three open-source models (GPT OSS 20B, Mistral Medium, and DeepSeek R1). Following each conversation, the expert curates a rubric that evaluates the final model response. We open-source all 1,100 conversations used in this paper for evaluation, while retaining a private heldout set to monitor potential data contamination in future model releases.

\begin{figure}[!ht]
    \centering
    \includegraphics[width=0.95\textwidth]{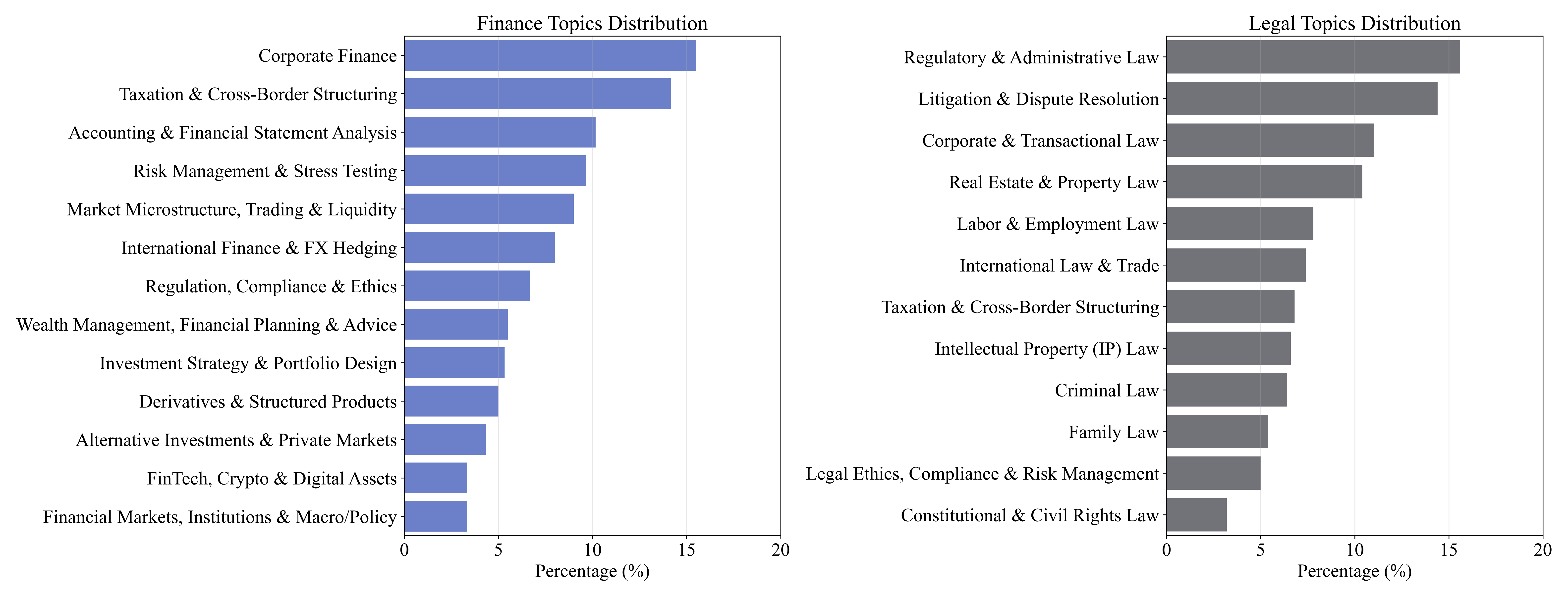}
    \caption{Topic distribution of prompts in \benchmarkname across Finance and Legal domains.}
    \label{fig:topic_distribution}
\end{figure}

The datasets span 13 Finance and 12 Legal topics, initially inspired by real usage data in Scale Showdown\footnote{\href{https://scale.com/showdown}{https://scale.com/showdown}} and subsequently refined in collaboration with domain experts to strike a balance between realism and difficulty. The resulting distribution is provided in \cref{fig:topic_distribution}. We automatically classify \benchmarkname conversations into jurisdictions and find that they span over 114 and 47 jurisdictions globally and in the US, respectively. Following \cite{arora2025healthbench}, we order the conversations by difficulty based on average scores across all models evaluated in this work and split the 250 and 300 most difficult tasks for Legal and Finance into a \textit{Hard} subset for frontier models, respectively.

\section{Data Collection} \label{sec:data}
\begin{figure}[!ht] 
    \centering
    \includegraphics[width=\linewidth]{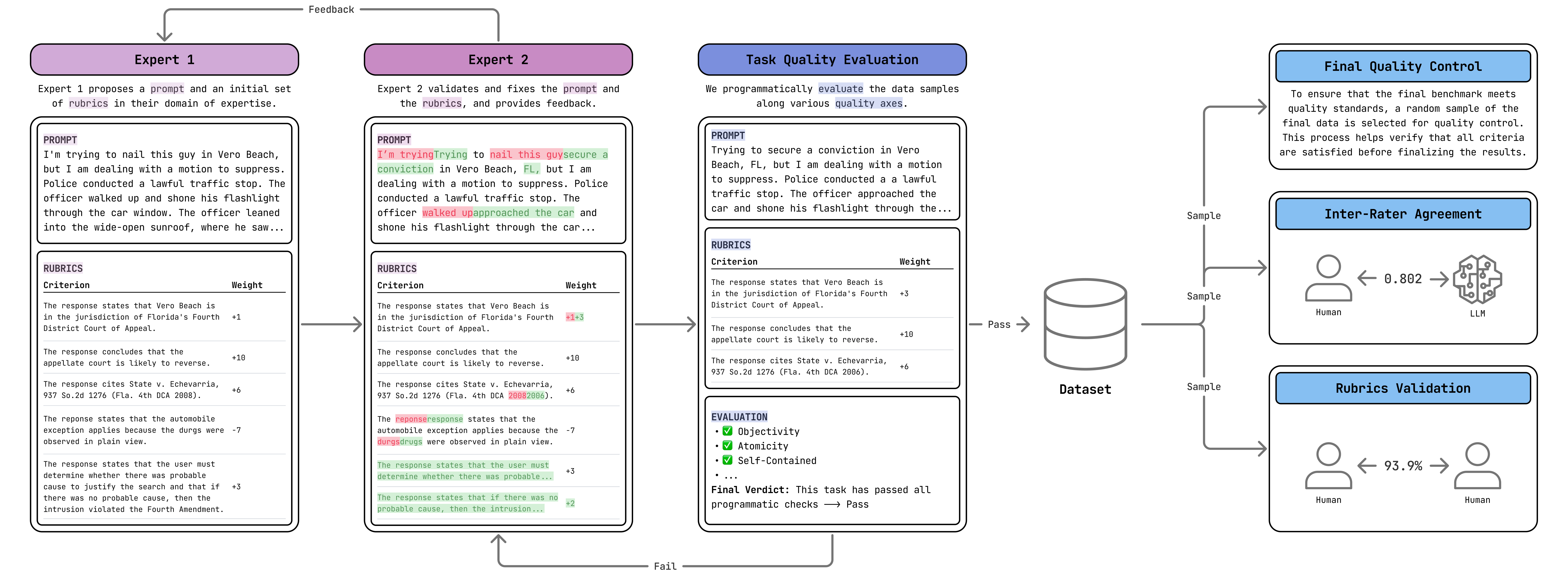}
    \caption{\textbf{Data Collection and Validation Pipeline}. Each prompt in the dataset is first authored by an annotator, either as a single-turn query or, in the case of multi-turn interactions, through a dialogue with a chat model. A second domain expert then reviews, edits, and provides feedback to the first author. To maintain objective, self-contained, and easy-to-grade criteria, we apply an automated validation procedure that checks for adherence to rubric design standards described in \cref{sec:data}.}
    \label{fig:data_collection}
\end{figure}

Among the 182 professionals, all annotators contributing to the Legal subset hold a JD or equivalent internationally and we require a Master's, CFA, or 6+ years of professional experience for Finance.

\subsection{Building Conversations}
Both benchmarks cover both expert and non-expert user questions. Participants are asked to contribute questions that either they or other experts in the field would actually care about, or those that they receive from their clients. Both types of questions, regardless, should require substantial analysis, interpretation, or creative thinking rather than just mechanical problem-solving. We discourage exam-like or theoretical questions that do not bear any real-life implications. Annotators only contribute to the topic that aligns well with their subject matter expertise.

\begin{figure}[t]
    \centering
    \includegraphics[width=1.0\textwidth]{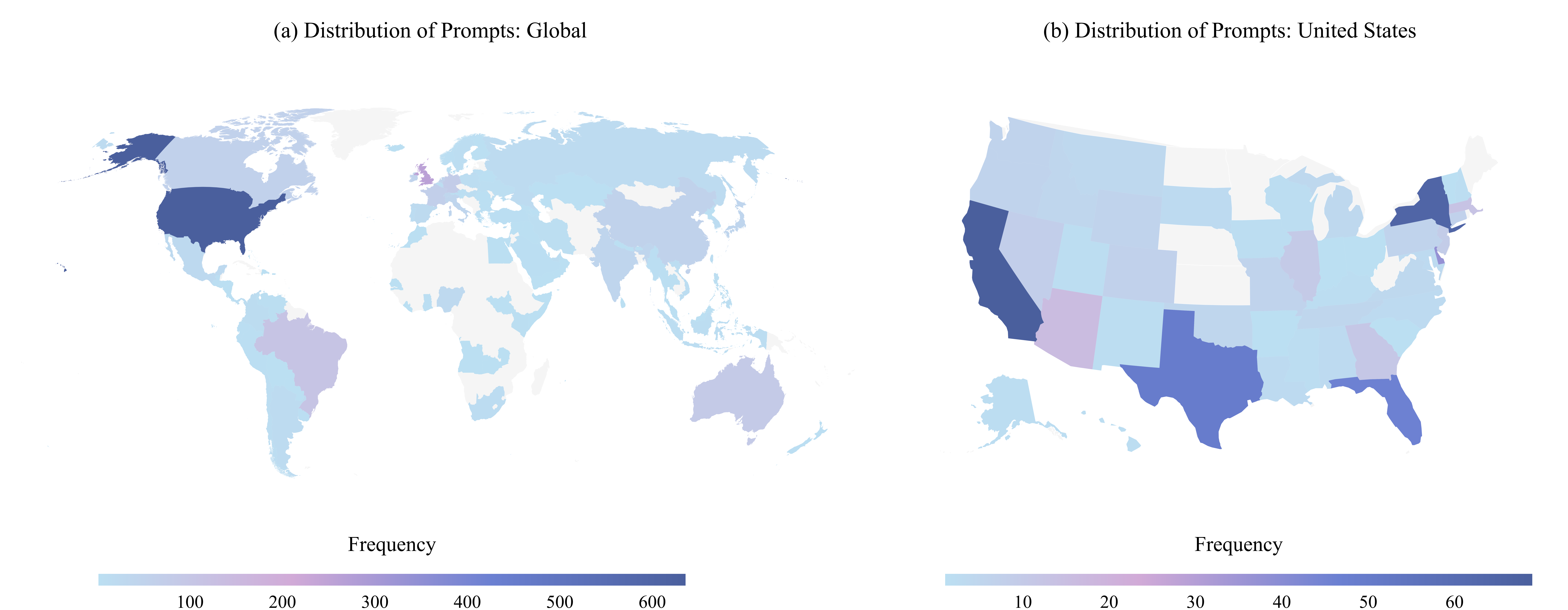}
    \caption{(a) A global frequency map showing the 114 countries and dependencies, and (b) A localized frequency map showing the 47 US states and territories covered by PRBench prompts across both Finance and Legal domains.}
    \label{fig:jurisdiction_heatmap}
\end{figure}

For about 30\% of the cases across Law and Finance (see \cref{tab:dataset_stats} for a full distribution), experts engage in a multi-turn conversation (up to 10 turns) with an open-source model to iteratively build context for the question or make clarifications (see \cref{dataset_details}). During this process, they are also encouraged to hint at relevant jurisdictions, if applicable, when curating their questions. This both enables the evaluation of jurisdiction-specific reasoning and helps reduce subjectivity in rubric creation.\footnote{This is also important in distinguishing between failure cases where a chat model has access to accurate location information (e.g., via memory or metadata) and those where it makes incorrect assumptions about the jurisdiction.} We automatically classify the prompts by jurisdiction and identify that they span 114 countries and dependencies spread globally. A further analysis of United States-specific prompts identifies 47 distinct jurisdictions (states and territories) represented across both Legal and Finance (see \cref{fig:jurisdiction_heatmap}). Finally, for a small set of the datasets, experts included a set of reference texts. Those are pre-pended to the respective user turns. 

\subsection{Creating Rubrics}
\label{sec:creating_rubrics}
For each prompt, experts created a set of criteria (referred to as \textit{rubrics}) with associated integer weights between -10 and 10, excluding 0. Each criterion with a positive score (\textit{positive criterion}) describes a desired quality, whereas a criterion with a negative score (\textit{negative criterion}) describes undesired properties of a good quality response, such as \textit{``The response discusses IRC Section 355''} when Section 355 is irrelevant to the prompt. The resulting distribution of the scores can be found in \cref{fig:weight_distributions}. Below is the desiderata for rubric creation that were followed by human experts and enforced by quality control layers:
\begin{itemize}
    \item \textbf{Constructive}: Each criterion should be correct, precise, and free of internal errors or misconceptions.
    \item \textbf{Mutually Exclusive and Collectively Exhaustive}: No criterion is repeated or redundant, so that a model is not penalized twice for the same mistake. At the same time, the sum of all criteria should be thorough enough to cover all aspects of a perfect response.
    \item \textbf{Atomic}: Each rubric criterion evaluates exactly one distinct aspect and should contain no bundling of multiple criteria into a single criterion.
    \item \textbf{Objective}: Criteria should be binary (true or false) and objective, where a majority of readers should agree on whether a given model response satisfies the criteria.
    \item \textbf{Self-Contained}: All info needed to score a response must be included in the criterion.
\end{itemize}

For each criterion, annotators select one of six severity levels, ranging from \textit{Critically Important} to \textit{Critically Detrimental}, before assigning a corresponding weight (see \cref{tab:rubric_scale}). This procedure encourages hierarchical reasoning and helps reduce noise and inconsistency in weight assignments.

\begin{table}[ht]
\caption{Rubric criteria scoring levels.}
\centering
\small
\begin{tabularx}{\linewidth}{cX}
\toprule
\textbf{Score Range} & \textbf{Description} \\
\midrule
$\cinterval{+8, +10}$ & \textbf{Critically Important}: These are essential criteria without which the response would fail to adequately address the prompt. They define the minimally viable rubric set and capture only the core, indispensable elements of a correct and sufficient answer. \\
$\cinterval{+4, +7}$ & \textbf{Important}: Criteria that meaningfully strengthen a response by adding depth, accuracy, or completeness. They materially shape the response's quality but are not strictly required for it to be acceptable. \\
$\cinterval{+1, +3}$ & \textbf{Slightly Important}: Optional enhancements or ``nice-to-have'' details that improve clarity or precision but do not affect the core correctness of the response. \\
$\cinterval{-3, -1}$ & \textbf{Slightly Detrimental}: Minor issues or irrelevant tangents that slightly detract from quality or focus but do not undermine reasoning or factual integrity. \\
$\cinterval{-7, -4}$ & \textbf{Detrimental}: Significant errors or omissions that meaningfully weaken the response, such as misleading reasoning, incorrect facts, or major structural flaws, though the response remains generally valid. \\
$\cinterval{-10, -8}$ & \textbf{Critically Detrimental}: Severe errors that render the response fundamentally invalid, harmful, or unethical. These issues directly contradict the prompt or destroy the credibility of the reasoning. \\
\bottomrule
\end{tabularx}
\label{tab:rubric_scale}
\end{table}

\paragraph{Rubric Categories} We work with domain experts and identify 7 and 8 distinctive axes for each criterion in the rubric for Finance and Legal domains, respectively. We identify 5 mutual categories across two domains: \textbf{Practical Utility}, \textbf{Handling Uncertainty}, \textbf{Supplemental Insight}, and \textbf{Instruction Following}. Moreover, we present Legal domain-specific criteria (\textbf{Legal Accuracy}, \textbf{Procedural Correctness and Risk \& Ethical Disclosure} and \textbf{Application of Law to the Facts} similar to \textit{Rule Application} from \citet{guha2023legalbench}) and three categories for Finance (\textbf{Financial Accuracy}, \textbf{Process Transparency \& Auditability}, and \textbf{Risk \& Regulatory Disclosure}). Collectively, these criteria along these aspects describe qualities of model responses. We define these rubric categories in further detail in Appendix \ref{sec:rubric_cat_defs}. Finally, rubric categories along with their respective model scores are listed in \cref{fig:results_per_rubric_category}; the frequency at which each rubric category appears is given in \cref{fig:rubric_categories}.

We depict our rubric creation and quality control framework in \cref{fig:data_collection} and further describe the rubric validation procedure in \cref{app:rubrics_validation}. In the quality control layer, a random subset of samples is manually reviewed for overall quality and correctness. At the end of data collection, an independent expert evaluates the final rubric itself, marking each criterion as agree or disagree to assess rubric clarity and validity. This step results in a \textbf{93.9\%} agreement between experts on the validity of rubrics.
\section{Evaluation}
\label{sec:results}
\subsection{Results}

\begin{figure}[h]
    \centering
    \includegraphics[width=0.95\textwidth]{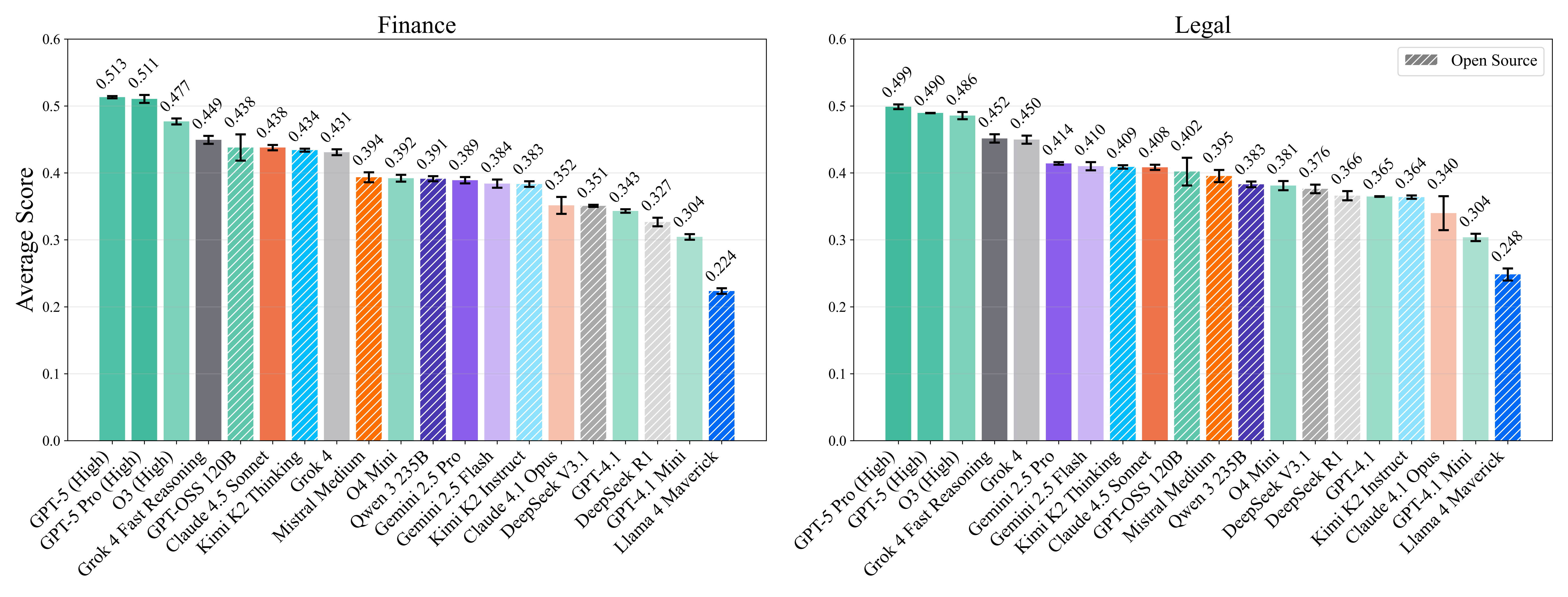}
    \caption{Results on all samples from the Legal and Finance domains of \benchmarkname. Each evaluation is repeated three times. We report the average and the 95\% confidence intervals for each model.}
    \label{fig:scores_all}
\end{figure}

\begin{figure}[t]
    \centering
    \includegraphics[width=0.95\textwidth]{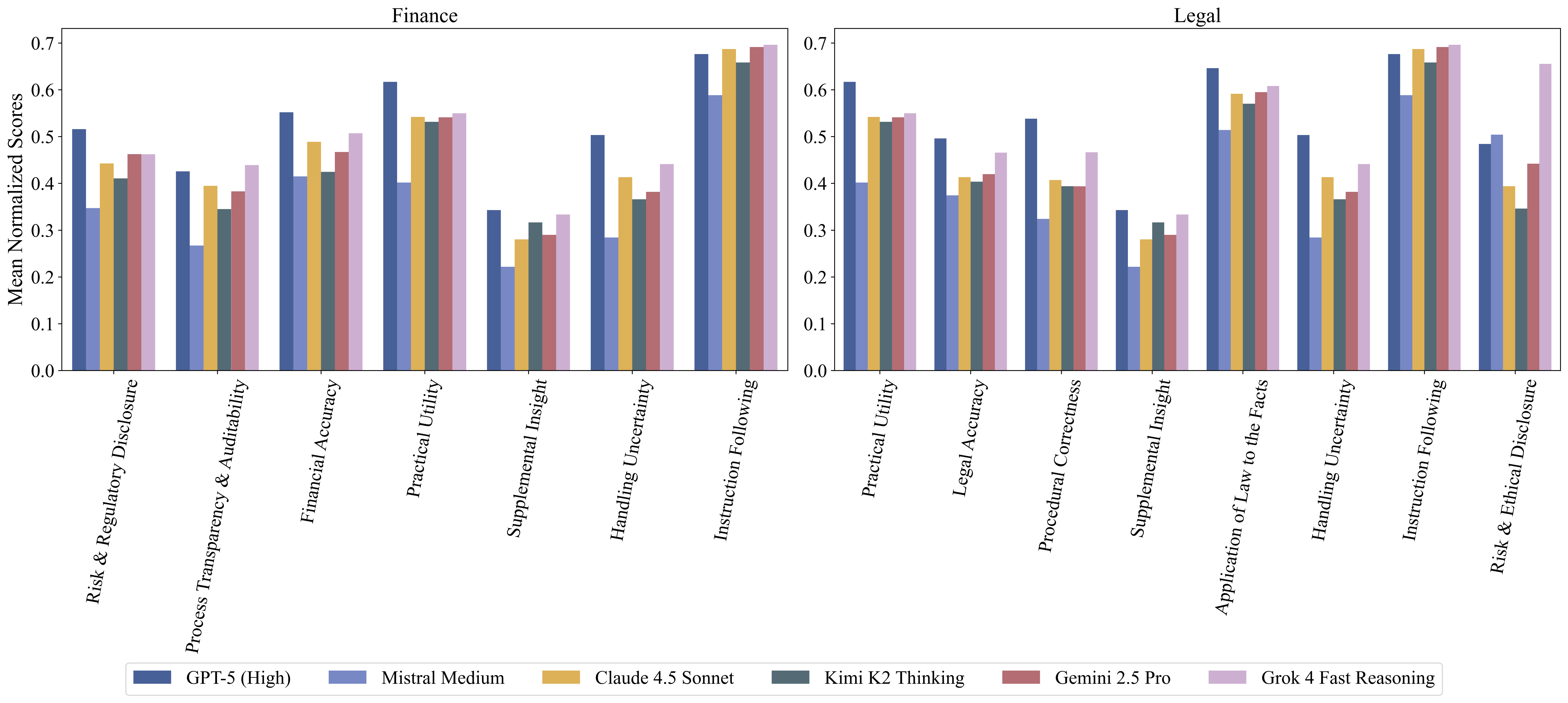}
    \caption{Min-normalized scores per rubric category for \benchmarkname. Further details on min-normalized scoring are provided in \cref{app:calculating_scores}.}
    \label{fig:results_per_rubric_category}
\end{figure}

We evaluate 20 different chat models using an LLM-based grader. Following \citet{arora2025healthbench}, we calculate the overall score by taking the mean of scores for each example and clipping it to be between $\cinterval{0,1}$ in \cref{fig:scores_all} and \cref{fig:hard_subsets}. In \cref{fig:results_per_rubric_category}, we normalize the individual model scores between the possible minimum and maximum scores (Min-Normalized Scoring). Further details on our scoring mechanism are provided in Appendix \ref{app:calculating_scores}.

We set the reasoning mode to \textit{High} for GPT-5, O3, and Grok-4 models. For Claude Sonnet 4.5 and Claude Opus 4.1, we set the thinking budget to 32K and 16K tokens, respectively. For Gemini 2.5 Pro, we experiment with setting a reasoning budget for 32K and dynamic thinking, achieving the best results with the dynamic configuration. We set the timeout to 60 minutes for every model and try for five attempts, and in the last attempt, we change the reasoning effort from high to low.

On the full set in \cref{fig:scores_all}, the top scores are 0.51 and 0.50 for Finance and Legal, respectively. On the Hard subset (see \cref{fig:hard_subsets}), the best-performing model achieves 0.39 and 0.37 for Finance and Legal, with one of GPT-5 and GPT-5 Pro consistently leading, followed by Grok 4 Fast Reasoning. The open-sourced models Kimi K2 Thinking and GPT OSS 120B closely follow the proprietary models. We additionally observe that more recent models (we use heavier color gradients for newer models) are able to consistently improve on this benchmark, indicating steady progress in professional reasoning capabilities.

\paragraph{Tool-Enabled Evaluations} The questions in both benchmarks are explicitly designed to be solvable through reasoning alone, without requiring external tools. Nevertheless, models often benefit from tool calls for case lookups or computations. The results for O3, GPT-5 and Grok 4 Fast Reasoning can be found in \cref{app:tool_evals}. Enabling web-search improves performance for O3 and Grok. We find that code interpreter offers no additional gains beyond those achieved using web search. All performances remain $\leq 0.4$ for the Hard subset with O3 leading. 

\paragraph{Performance across Rubric Categories} In \cref{fig:results_per_rubric_category}, we examine the performances of the top six models within each rubric category. For this analysis, we use an alternative scoring mechanism which we call \textit{min-normalized scores} as described in \cref{app:calculating_scores}. This metric is particularly suitable for category-level comparisons, since for certain sample--category combinations, all criteria may be negative, in which case the default scoring method used by \citet{arora2025healthbench} fails to differentiate between models.

Although not among the top four models overall, Gemini 2.5 Pro performs remarkably well on \textit{Instruction Following} in both domains. Grok-4 Fast Reasoning leads by a large margin in \textit{Risk \& Ethical Disclosure} for the Legal domain, followed by Mistral Medium. GPT-5 achieves the highest scores in \textit{Handling Uncertainty} and \textit{Practical Utility} across both domains; in Legal specifically, it outperforms Grok-4 Fast Reasoning by more than 10\% in \textit{Procedural Correctness}. Finally, GPT-5 leads the \textit{Legal Accuracy} and \textit{Financial Accuracy} categories, followed by Grok 4 Fast Reasoning. Kimi K2 Thinking ranks among the top three models in \textit{Supplemental Insight}, though it does not stand out as distinctly strong in any particular category.

\begin{figure}[t]
    \centering
    \includegraphics[width=0.95\textwidth]{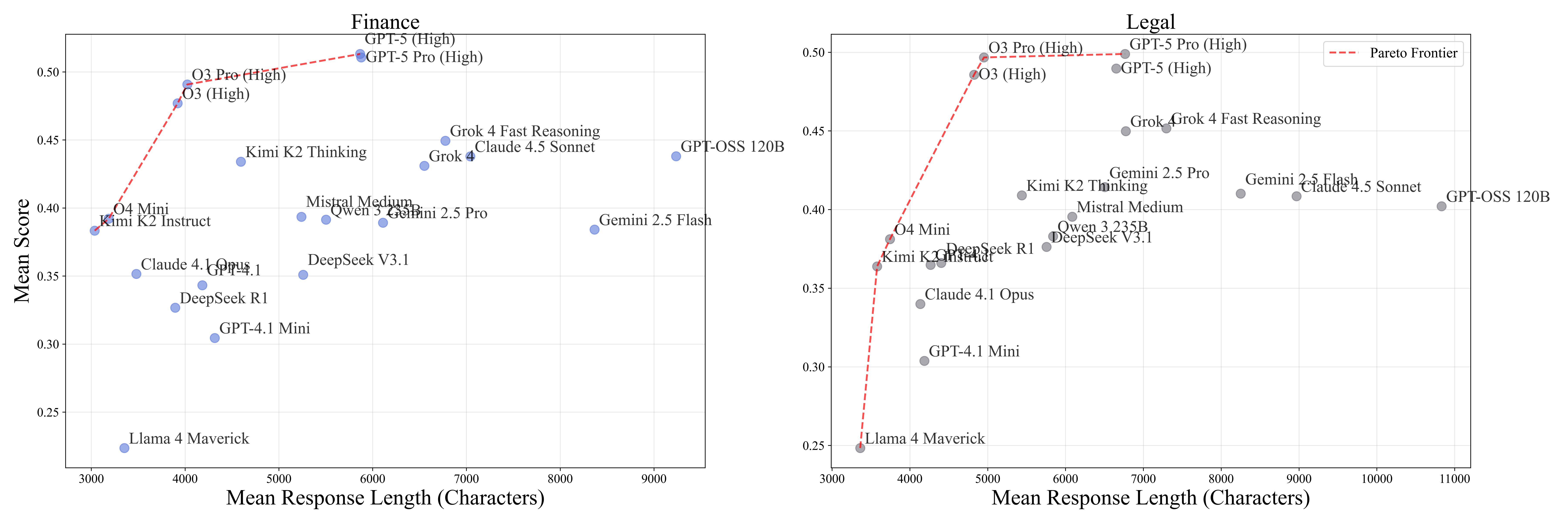}
    \caption{Scores vs.\ response lengths. While some models achieve similar performance (e.g., Kimi K2 Thinking and Claude Sonnet 4.5), conciseness appears as a differentiating factor. We only count alphanumeric characters, as most outputs contain Markdown formatting.}
    \label{fig:scores_vs_response_length}
\end{figure}

\paragraph{Controlling for Response Length} In \cref{fig:scores_vs_response_length}, we present average response lengths for all models. We observe that while Kimi K2 Thinking, Claude Sonnet 4.5, GPT OSS 120B, and Grok 4 Fast Reasoning achieve similar scores for Finance, Kimi K2 Thinking produces substantially shorter responses than the others. GPT-5 and O3 models strike a strong balance between performance and efficiency. In contrast, GPT-OSS 120B, Gemini 2.5 Flash, and Claude Sonnet 4.5 are the top three models that produce significantly longer responses, nearly twice the length of the best responses. 

\subsection{Evaluating LLM Judge}
\begin{table}[!htbp]
\centering
\begin{tabular}{lcc}
\toprule
 & \multicolumn{2}{c}{\textbf{Agreement with Experts}} \\
\cmidrule(lr){2-3}
\textbf{Judge} & Avg. Cohen's $\kappa$ & Avg. Macro F1 \\
\midrule
o4 Mini        & 0.603 & 0.801\\
GPT-4.1        & 0.605 & 0.802\\
Claude Haiku 4.5      & 0.535 & 0.765\\
Expert         & 0.589 & 0.813 \\
\bottomrule
\end{tabular}
\caption{LLM-Experts and Expert-Expert agreement for grading model responses over 101 tasks. For LLM judges, we calculate the average of agreement with two experts. For the expert judge, the agreement is calculated with the other expert. Agreements between a human and LLM judges are on par with the agreement between two experts, except for Claude Haiku 4.5.} 
\label{tab:judge_metrics_two_experts}
\end{table}

To determine which model to use as a judge, we measured the inter-rater agreement between model and expert labels over a collection of 101 tasks. More specifically, we had each judge and two experts grade GPT-5 and Claude Sonnet 4.5 responses for each task, indicating whether each rubric criterion was present or not. All grades were then pooled together, and we measure the average Cohen's $\kappa$ and Macro F1 scores between each judge and both humans. We report these scores in \cref{tab:judge_metrics_two_experts}. We note that our LLM-expert and expert-expert Macro F1 scores exceed the scores reported in previous work \citep{arora2025healthbench}.

We find that all judges demonstrate high and similar agreement scores with both experts, with the exception of Claude Haiku 4.5. In light of these comparable IRA scores, we select \textit{o4-mini} due to its reduced querying costs.

\section{Dissecting the Dataset: Prompts, Rubrics, and Beyond}

\subsection{Inside the Conversations}
\begin{figure}[t]
    \centering
    \includegraphics[width=0.95\textwidth]{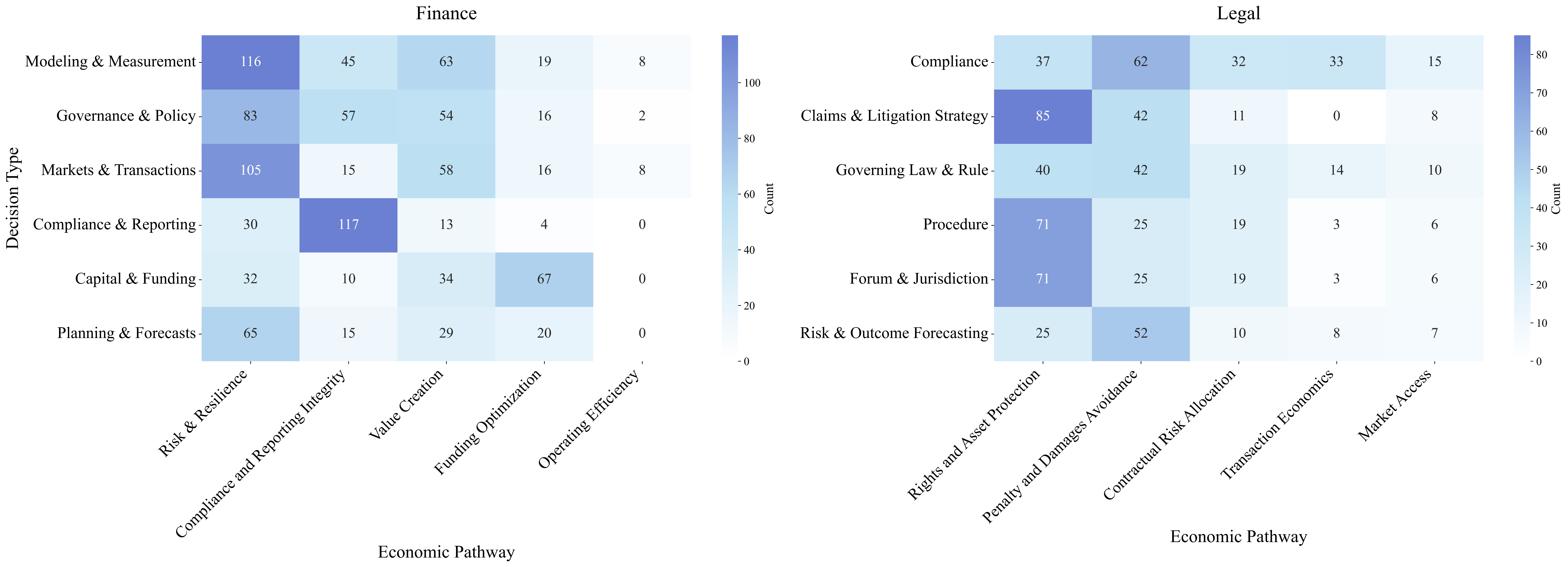}
    \caption{Distribution of \textit{Decision Types} and \textit{Economic Pathways} across Legal and Finance domains.}
    \label{fig:decisionvseconomic}
\end{figure}

Beyond the existing topic distribution, we analyze the capabilities tested by our benchmarks along two additional axes: the type of decision the question seeks assistance with, and the economic implications it entails. Specifically, these axes address the questions \textit{``What kind of decision is being made?''} and \textit{``What economic consequence does it affect?''}. We refer to the former as the \textit{Decision Type} and the latter as the \textit{Economic Pathway}. We name the latter category to capture the idea of tracing \textit{pathways of value, risk, or cost}. Overall, we find that the majority of questions in our dataset correspond to high-stakes, real-world decision scenarios that also imply tangible downstream economic impact. The resulting distribution for these dimensions is given in \cref{fig:decisionvseconomic}. All annotations are released alongside the dataset to facilitate future research.

\subsubsection{Is AI Ready for Assisting High-Stakes Decisions?}
AI systems are increasingly being deployed to support human decision-making across domains such as law, finance, healthcare, and management~\citep{Zeiser2024,Kim2025,Khosravi2024,Vukovic2025,Hillebrand2025}. Yet, evidence on their effectiveness remains mixed: while some studies find that AI assistance can improve consistency and reduce cognitive load, others show that it can amplify errors when models provide incorrect or oversimplified recommendations~\citep{rojas2024gazette,steyvers2024three,eigner2024determinants}. Within our dataset, many prompts extend beyond factual or informational queries, posing genuine decision problems, such as whether to litigate or arbitrate, how to allocate funds, or which market to launch in. To better understand how LLMs engage with such high-stakes reasoning tasks, we recruit domain experts to annotate each sample with the type of decision it represents. This enables future research to systematically analyze which categories of decisions current models handle well or poorly, providing a foundation for more targeted evaluations of AI-assisted decision-making. 

\subsubsection{Can AI Handle Economically Consequential Questions?}

Recent benchmark efforts have shifted focus from academic problem-solving toward economically valuable tasks that reflect how AI systems can drive productivity and create measurable real-world value \citep{gdpval,rli}. In our dataset, a majority of prompts naturally carry downstream economic implications: for instance, advising how to allocate capital expenditures under interest-rate shocks, whether to diversify or concentrate portfolio exposure, how to structure cross-border acquisitions to minimize risk, or how to design stress tests that prevent catastrophic losses. In these scenarios, a model's performance can meaningfully affect financial outcomes—saving or losing money, reducing risk, or improving operational efficiency. To capture this dimension, we recruit domain experts to annotate each sample for its economic pathway, indicating whether and how the question implies a positive or negative economic consequence if answered correctly or incorrectly.

\subsection{Inside the Rubrics}

\begin{figure}[!ht]
    \centering
    \includegraphics[width=\linewidth]{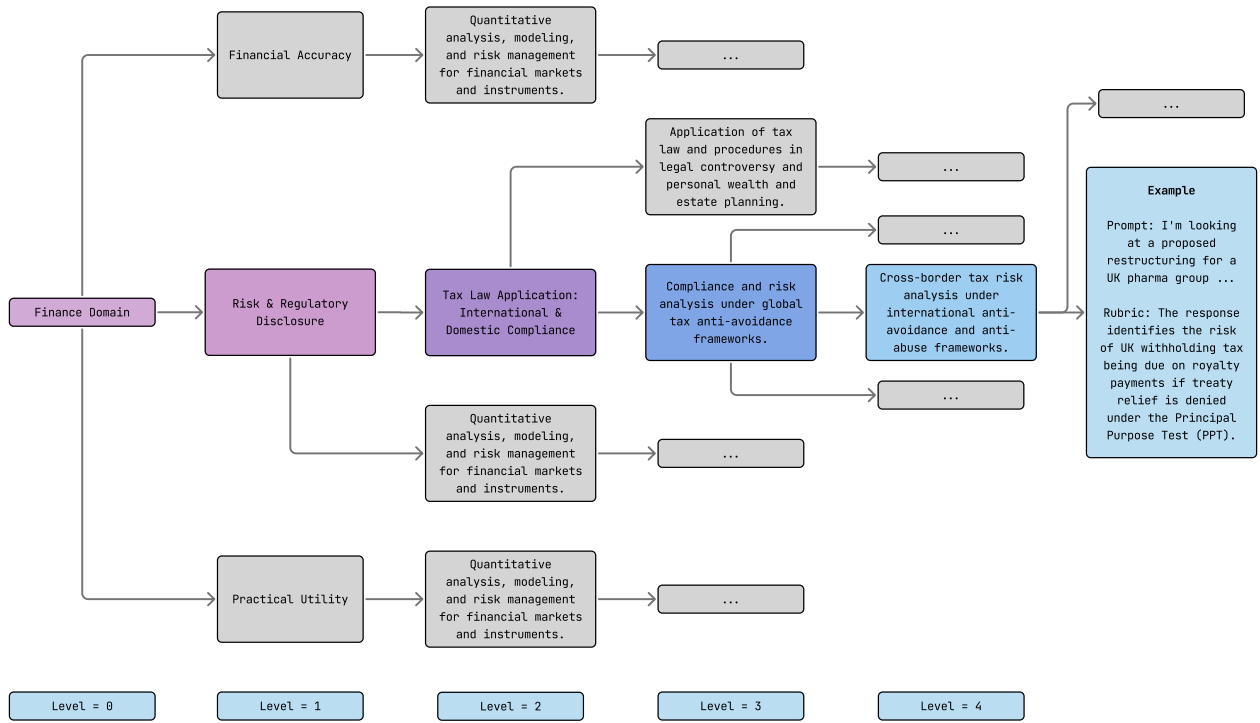}
    \caption{Automatically clustering the available rubric criteria enables fine-grained analysis of failure modes.}
    \label{fig:cluster}
\end{figure}

To perform a granular analysis on rubrics, we perform hierarchical clustering on rubrics. The objective is to move from high-level performance scores to a fine-grained understanding of specific model capability gaps. We first identify the capabilities required to score highly on each rubric and then we construct a five-level hierarchical clustering tree by clubbing similar capabilities together. 

We perform this analysis independently for the Finance and Legal domains. Using the model performance of each rubric, we identify fine-grained clusters where the model demonstrates significant underperformance compared to other clusters. This information can be used to guide further finetuning or further analysis of model behavior.

\cref{fig:cluster} illustrates the hierarchical clustering using a representative example from out dataset from the Finance domain. The hierarchy includes four levels of clustering. In this instance, we analyze the criteria from the ``Risk and Regulatory Disclosure'' rubric category (Level 1 in \cref{fig:cluster}). We then dynamically derive the subsequent granular layers (Levels 2 to 4) across the entire dataset. Specifically, this example maps to ``The Law Application: International and Domestic Compliance'' at Level 2 and further refines to ``Compliance and risk analysis under global tax anti-avoidance frameworks'' at Level 3. This multi-level hierarchical clustering enables flexibility in deriving insights, allowing us to select a specific level of granularity depending on interested capabilities.

Applying this methodology to our rubrics shows significant performance differences between models on capability clusters. For instance, within the Legal domain, the cluster ``Advanced corporate and \textit{international tax law}, strategy, and compliance services.'', shows a significant performance difference. GPT-5 achieved 0.64 accuracy, whereas Grok-4 only achieved 0.16. A similar disparity was observed in the Finance domain for ``Strategic planning, compliance, and optimization for international corporate tax.'' cluster. Here, we found that Claude Opus 4.1 lags behind substantially, with only 0.34, where the top performing model achieved 0.76.

\section{Related Work}
\label{sec:relatedwork}

\begin{table*}[!htbp]
\centering
\small
\caption{Comparison of \benchmarkname with select professional-domain benchmarks. GT stands for Ground-Truth based evaluation. \textit{Partial} indicates a non-major subset of the dataset involved Open-Ended QA.}
\label{tab:benchmark_comparison}
\resizebox{\textwidth}{!}{
\begin{tabular}{l r c l r c c c}
\toprule
\textbf{Benchmark} & \textbf{\# Samples} & \textbf{Open-Ended QA} & \textbf{Evaluation} & \textbf{\# Rubrics} & \textbf{Multi-Domain} & \textbf{Multi-Turn} & \textbf{Open-Source} \\
\midrule
LegalBench \citep{guha2023legalbench}        & 162    & \textit{Partial}       & GT                 &    --        & {\color{red}\ding{55}} & {\color{red}\ding{55}} & {\color{ao}\ding{51}} \\
LEXam \citep{Fan2025LEXamBL}             & 4,886  & \textit{Partial}       & GT, LLM Judge      &    --        & {\color{red}\ding{55}} & {\color{red}\ding{55}} & {\color{ao}\ding{51}} \\
BigLawBench \citep{biglaw}       & \textit{Private} & {\color{ao}\ding{51}} & Rubric, LLM Judge  & \textit{Private} & {\color{red}\ding{55}} & {\color{red}\ding{55}} & {\color{red}\ding{55}} \\
\midrule
CorpFin v2 \citep{ValsAI2025CorpFinV2}        & 858    & {\color{ao}\ding{51}} & GT, LLM Judge      &    --        & {\color{red}\ding{55}} & {\color{red}\ding{55}} & {\color{red}\ding{55}} \\
ConvFinQA \citep{Chen2022ConvFinQAET}         & 8,281  & {\color{red}\ding{55}} & GT                 &    --        & {\color{red}\ding{55}} & {\color{ao}\ding{51}} & {\color{ao}\ding{51}} \\
FinanceBench \citep{Islam2023FinanceBenchAN}      & 10,231 & {\color{red}\ding{55}} & GT                 &    --        & {\color{red}\ding{55}} & {\color{red}\ding{55}} & {\color{ao}\ding{51}} \\
\midrule
ProfBench \citep{profbench}         & 80     & {\color{ao}\ding{51}} & Rubric, LLM Judge  & 2,448        & {\color{ao}\ding{51}} & {\color{red}\ding{55}} & {\color{ao}\ding{51}} \\
APEX \citep{apex}              & 200    & {\color{ao}\ding{51}} & Rubric, LLM Judge  & 5,818        & {\color{ao}\ding{51}} & {\color{red}\ding{55}} & {\color{red}\ding{55}} \\
HealthBench \citep{arora2025healthbench}       & 5,000  & {\color{ao}\ding{51}} & Rubric, LLM Judge  & 48,562       & {\color{red}\ding{55}} & {\color{ao}\ding{51}} & {\color{ao}\ding{51}} \\
\midrule
\textsc{PRBench} (Ours)       & 1,100  & {\color{ao}\ding{51}} & Rubric, LLM Judge  & 19,356       & {\color{ao}\ding{51}} & {\color{ao}\ding{51}} & {\color{ao}\ding{51}} \\
\bottomrule
\end{tabular}
}
\vspace{-10pt}
\end{table*}

\subsection{Evaluating LLMs on Economically Valuable Tasks}
While dominant expert-level knowledge benchmarks like MMLU~\citep{hendrycks2021mmlu} and GPQA~\citep{rein2024gpqa} test academic reasoning, a recent trend focuses on evaluating AI performance on professional and economically valuable tasks. This includes benchmarks like SWE-Lancer~\citep{swelancer} for freelance software engineering, GDPval~\citep{gdpval} for tasks across U.S. GDP-contributing occupations, APEX~\citep{apex} for high-value work in consulting, finance, law, and healthcare, ProfBench~\citep{profbench} that covers tasks in finance and consulting, HealthBench~\citep{arora2025healthbench} focusing solely on health care, and BigLaw Bench~\citep{biglaw} for complex legal tasks. Other novel approaches include the Remote Labor Index (RLI)~\citep{rli}, which measures AI's automation potential for remote work, AlphaArena~\citep{AlphaArena2025}, which evaluates AI agents in live financial trading competitions, and Arena Expert\citep{arenaexpert}, which uses human preference voting to evaluate occupational tasks.

However, existing evals for open-ended professional tasks are often limited. They tend to be private~\citep{apex, gdpval, biglaw} or require costly human expert judges~\citep{arenaexpert, rli}, limiting accessibility and scalability. Furthermore, potentially due to the cost of sourcing expert annotations, existing rubric-based professional benchmarks are often small in scale~\citep{apex, profbench, biglaw}, lacking sufficient coverage of diverse professional topics.

Our work, \benchmarkname, offers a significant public set of 1,100 tasks and 19,356 expert-curated criteria that enables self-served evaluation, an order of magnitude larger than benchmarks like APEX~\citep{apex}, ProfBench~\citep{profbench}, and Biglaw Bench~\citep{biglaw}, and uniquely leverages multi-turn interactions to build up the context of real legal and finance settings.

\subsection{Rubric-Based Evaluation and Reward} Rubric-based evaluation is a key methodology for enabling scalable, automated evaluation of open-ended professional tasks. Its use of self-contained, objective criteria provides the objectivity and style-neutrality essential for knowledge-intensive or reasoning tasks. This approach is used in various recent benchmarks, from general-domain evaluations~\citep{multichallenge, ifeval} to professional domain evaluations~\citep{profbench, apex, arora2025healthbench}. It is also applied in agent-focused benchmarks like RLI~\citep{rli} and BrowseComp \citep{browsecomp} to grade complex task artifacts. Beyond evaluation, rubrics are also being explored as a reward function for reinforcement learning~\citep{rar, rubricsanchor}. Further studies have explored rubric synthesis techniques from diverse responses~\citep{chasingthetail, onlinerubrics}.

Our work, \benchmarkname, expands this line of research by contributing a large-scale, public resource of expert-curated criteria, which are annotated with detailed categories across legal and finance to enable future research on professional reasoning and reward modeling.
\section{Conclusion}
\label{sec:conclusion}

We introduced \benchmarkname, a large-scale expert-annotated benchmark for evaluating LLMs on high-stakes professional reasoning in Finance and Law, two domains where reasoning quality directly affects real-world outcomes. By combining over 1,100 expert-authored tasks and 19,000+ rubric criteria, \benchmarkname enables interpretable, rubric-based evaluation of models on open-ended, economically consequential problems. Our analysis shows that while both proprietary and open-source models demonstrate steady progress, substantial gaps remain in process transparency and domain-specific diligence. Models frequently reach conclusions through incomplete or opaque reasoning, limiting their trustworthiness in professional settings. \benchmarkname provides a framework for objective, fine-grained evaluation of model reasoning. By making this benchmark publicly available, we aim to advance research toward transparent, reliable, and economically valuable AI systems capable of assisting in real-world decision-making.

\section*{Acknowledgments}
We thank Karmini Sampath, Emily Chan, Neel Guha and Jerry Xu for helpful feedback and discussions during the development of this work. We also thank Amir Fekrazad, Gabrial Mathews, Shannon Blakeney, Jermaine Ogwuda, Valerie Muigai, Karen Knighton, Diana Bonilla, and Hayden Morse for their contributions to data validation and quality control.

\bibliography{custom}

@article{chollet2025arc,
  title={Arc-agi-2: A new challenge for frontier ai reasoning systems},
  author={Chollet, Francois and Knoop, Mike and Kamradt, Gregory and Landers, Bryan and Pinkard, Henry},
  journal={arXiv preprint arXiv:2505.11831},
  year={2025}
}

@techreport{scale_showdown_2025,
  title        = {SEAL Showdown: Technical Report},
  author       = {{Scale AI}},
  institution  = {Scale AI},
  year         = {2025},
  month        = {September},
  url          = {https://showdown.scale.com/assets/SEAL_Showdown_Tech_Report.pdf},
  note         = {Preliminary results and methodology for the SEAL Showdown leaderboard}
}

@article{appel2025anthropic,
  title={Anthropic economic index report: Uneven geographic and enterprise ai adoption},
  author={Appel, Ruth and McCrory, Peter and Tamkin, Alex and Stern, Michael and McCain, Miles and Neylon, Tyler},
  journal={Anthropic Research},
  year={2025}
}

@misc{rli,
      title={Remote Labor Index: Measuring AI Automation of Remote Work}, 
      author={Mantas Mazeika and Alice Gatti and Cristina Menghini and Udari Madhushani Sehwag and Shivam Singhal and Yury Orlovskiy and Steven Basart and Manasi Sharma and Denis Peskoff and Elaine Lau and Jaehyuk Lim and Lachlan Carroll and Alice Blair and Vinaya Sivakumar and Sumana Basu and Brad Kenstler and Yuntao Ma and Julian Michael and Xiaoke Li and Oliver Ingebretsen and Aditya Mehta and Jean Mottola and John Teichmann and Kevin Yu and Zaina Shaik and Adam Khoja and Richard Ren and Jason Hausenloy and Long Phan and Ye Htet and Ankit Aich and Tahseen Rabbani and Vivswan Shah and Andriy Novykov and Felix Binder and Kirill Chugunov and Luis Ramirez and Matias Geralnik and Hernán Mesura and Dean Lee and Ed-Yeremai Hernandez Cardona and Annette Diamond and Summer Yue and Alexandr Wang and Bing Liu and Ernesto Hernandez and Dan Hendrycks},
      year={2025},
      eprint={2510.26787},
      archivePrefix={arXiv},
      primaryClass={cs.LG},
      url={https://arxiv.org/abs/2510.26787}, 
}

@techreport{chatterji2025people,
  title={How people use chatgpt},
  author={Chatterji, Aaron and Cunningham, Thomas and Deming, David J and Hitzig, Zoe and Ong, Christopher and Shan, Carl Yan and Wadman, Kevin},
  year={2025},
  institution={National Bureau of Economic Research}
}

@article{LMArena2025ArenaExpert,
  author    = {LMArena Team},
  title     = {Arena Expert and Occupational Categories},
  journal   = {LMArena News},
  year      = {2025},
  month     = nov,
  day       = {05},
  url       = {https://news.lmarena.ai/arena-expert/},
  note      = {Blog post}
}

@misc{apex,
      title={The AI Productivity Index (APEX)}, 
      author={Bertie Vidgen and Abby Fennelly and Evan Pinnix and Chirag Mahapatra and Zach Richards and Austin Bridges and Calix Huang and Ben Hunsberger and Fez Zafar and Brendan Foody and Dominic Barton and Cass R. Sunstein and Eric Topol and Osvald Nitski},
      year={2025},
      eprint={2509.25721},
      archivePrefix={arXiv},
      primaryClass={econ.GN},
      url={https://arxiv.org/abs/2509.25721}, 
}

@misc{gdpval,
      title={GDPval: Evaluating AI Model Performance on Real-World Economically Valuable Tasks}, 
      author={Tejal Patwardhan and Rachel Dias and Elizabeth Proehl and Grace Kim and Michele Wang and Olivia Watkins and Simón Posada Fishman and Marwan Aljubeh and Phoebe Thacker and Laurance Fauconnet and Natalie S. Kim and Patrick Chao and Samuel Miserendino and Gildas Chabot and David Li and Michael Sharman and Alexandra Barr and Amelia Glaese and Jerry Tworek},
      year={2025},
      eprint={2510.04374},
      archivePrefix={arXiv},
      primaryClass={cs.LG},
      url={https://arxiv.org/abs/2510.04374}, 
}

@misc{swelancer,
  title  = {SWE-Lancer: Can Frontier LLMs Earn \$1 Million from Real-World Freelance Software Engineering?},
  author = {Samuel Miserendino and Michele Wang and Tejal Patwardhan and Johannes Heidecke},
  year   = {2025},
  eprint = {2502.12115},
  archivePrefix = {arXiv},
  primaryClass  = {cs.LG},
  url    = {https://arxiv.org/abs/2502.12115}
}

@misc{AlphaArena2025,
  title        = {Alpha Arena},
  author       = {nof1.ai},
  year         = {2025},
  howpublished = {\url{https://nof1.ai/}},
  note         = {Accessed: 2025-11-11}
}

@misc{biglaw,
  author = {Pereyra, Julio and Lebens, Elizabeth and Guillod, Matthew and Toulme, Laura and MacGregor, Cameron and Murdter, David and de la Roche, Karl and McConnachie, Emilie and Pushkin, Jeremy and Kim, Rina and Chan, Aaron and Pan, Jenny and Yang, Boling and Wu, Nan and Grupen, Niko and Oh, Lauren and Nayak, Aatish and Pereyra, Gabriel},
  title = {Introducing BigLaw Bench},
  year = {2024},
  month = {Aug},
  day = {29},
  howpublished = {\url{https://www.harvey.ai/blog/introducing-biglaw-bench}},
  note = {Accessed: 2025-11-11}
}

@misc{arenaexpert,
  author = {{LMArena Team}},
  title = {Arena Expert and Occupational Categories},
  year = {2025},
  month = {Nov},
  day = {5},
  howpublished = {\url{https://news.lmarena.ai/arena-expert/}},
  note = {Accessed: 2025-11-11}
}

@article{arora2025healthbench,
  title={Healthbench: Evaluating large language models towards improved human health},
  author={Arora, Rahul K and Wei, Jason and Hicks, Rebecca Soskin and Bowman, Preston and Qui{\~n}onero-Candela, Joaquin and Tsimpourlas, Foivos and Sharman, Michael and Shah, Meghan and Vallone, Andrea and Beutel, Alex and others},
  journal={arXiv preprint arXiv:2505.08775},
  year={2025}
}

@inproceedings{
guha2023legalbench,
title={LegalBench: A Collaboratively Built Benchmark for Measuring Legal Reasoning in Large Language Models},
author={Neel Guha and Julian Nyarko and Daniel E. Ho and Christopher Re and Adam Chilton and Aditya Narayana and Alex Chohlas-Wood and Austin Peters and Brandon Waldon and Daniel Rockmore and Diego Zambrano and Dmitry Talisman and Enam Hoque and Faiz Surani and Frank Fagan and Galit Sarfaty and Gregory M. Dickinson and Haggai Porat and Jason Hegland and Jessica Wu and Joe Nudell and Joel Niklaus and John J Nay and Jonathan H. Choi and Kevin Tobia and Margaret Hagan and Megan Ma and Michael Livermore and Nikon Rasumov-Rahe and Nils Holzenberger and Noam Kolt and Peter Henderson and Sean Rehaag and Sharad Goel and Shang Gao and Spencer Williams and Sunny Gandhi and Tom Zur and Varun Iyer and Zehua Li},
booktitle={Thirty-seventh Conference on Neural Information Processing Systems Datasets and Benchmarks Track},
year={2023},
url={https://openreview.net/forum?id=WqSPQFxFRC}
}

@article{eigner2024determinants,
  title   = {Determinants of LLM-assisted Decision-Making},
  author  = {Eigner, Eva and H{\"a}ndler, Thorsten},
  journal = {arXiv preprint arXiv:2402.17385},
  year    = {2024},
  url     = {https://arxiv.org/abs/2402.17385}
}

@article{steyvers2024three,
  title     = {Three Challenges for AI-Assisted Decision-Making},
  author    = {Steyvers, Mark and Kumar, Aakriti},
  journal   = {Perspectives on Psychological Science},
  volume    = {19},
  number    = {5},
  pages     = {722--734},
  year      = {2024},
  doi       = {10.1177/17456916231181102},
  url       = {https://pmc.ncbi.nlm.nih.gov/articles/PMC11373149/}
}

@online{rojas2024gazette,
  author    = {Rojas, Nikki},
  title     = {Does AI help humans make better decisions?},
  year      = {2024},
  month     = {June},
  day       = {14},
  url       = {https://news.harvard.edu/gazette/story/2024/06/does-ai-help-humans-make-better-decisions-artificial-intelligence-law/},
  note      = {Harvard Gazette}
}

@article{Fan2025LEXamBL,
  title={LEXam: Benchmarking Legal Reasoning on 340 Law Exams},
  author={Yu Fan and Jingwei Ni and Jakob Merane and Etienne Salimbeni and Yang Tian and Yoan Hermstruwer and Yinya Huang and Mubashara Akhtar and Florian Geering and Oliver Dreyer and Daniel Brunner and Markus Leippold and Mrinmaya Sachan and Alexander Stremitzer and Christoph Engel and Elliott Ash and Joel Niklaus},
  journal={ArXiv},
  year={2025},
  volume={abs/2505.12864},
  url={https://api.semanticscholar.org/CorpusID:278740278}
}

@inproceedings{Chen2022ConvFinQAET,
  title={ConvFinQA: Exploring the Chain of Numerical Reasoning in Conversational Finance Question Answering},
  author={Zhiyu Chen and SHIYANG LI and Charese Smiley and Zhiqiang Ma and Sameena Shah and William Yang Wang},
  booktitle={Conference on Empirical Methods in Natural Language Processing},
  year={2022},
  url={https://api.semanticscholar.org/CorpusID:252780839}
}

@article{Islam2023FinanceBenchAN,
  title={FinanceBench: A New Benchmark for Financial Question Answering},
  author={Pranab Islam and Anand Kannappan and Douwe Kiela and Rebecca Qian and Nino Scherrer and Bertie Vidgen},
  journal={ArXiv},
  year={2023},
  volume={abs/2311.11944},
  url={https://api.semanticscholar.org/CorpusID:265294665}
}

@misc{ValsAI2025CorpFinV2,
  author       = {Vals AI, Inc.},
  title        = {CorpFin (v2): A private benchmark evaluating understanding of long-context credit agreements},
  howpublished = {Online benchmark},
  year         = {2025},
  month        = nov,
  day          = {07},
  url          = {https://www.vals.ai/benchmarks/corp_fin_v2},
  note         = {Accessed on \today}
}

@article{Zeiser2024,
  author    = {Jannik Zeiser},
  title     = {Owning Decisions: AI Decision-Support and the Attributability-Gap},
  journal   = {Science and Engineering Ethics},
  volume    = {30},
  number    = {4},
  pages     = {1--19},
  year      = {2024},
  doi       = {10.1007/s11948-024-00485-1}
}

@article{Kim2025,
  author    = {Sihyun Kim and Sangyoon Yi and Sung{-}Pil Park},
  title     = {Prioritizing challenges in AI adoption for the legal domain: A systematic review and expert-driven {AHP} analysis},
  journal   = {PLOS ONE},
  volume    = {20},
  number    = {6},
  pages     = {e0326028},
  year      = {2025},
  doi       = {10.1371/journal.pone.0326028}
}

@article{Khosravi2024,
  author    = {Mohsen Khosravi and Zahra Zare and Seyyed M. Mojtabaeian and Reyhane Izadi},
  title     = {Artificial Intelligence and Decision-Making in Healthcare: A Thematic Analysis of a Systematic Review of Reviews},
  journal   = {Health Services Research and Managerial Epidemiology},
  volume    = {11},
}

@article{Vukovic2025,
  author    = {Darko B. Vukovi{\'c} and Senanu Dekpo{-}Adza and Stefana Matovi{\'c}},
  title     = {{AI} integration in financial services: A systematic review of trends and regulatory challenges},
  journal   = {Humanities and Social Sciences Communications},
  volume    = {12},
  number    = {1},
  pages     = {562},
  year      = {2025},
  doi       = {10.1057/s41599-025-04850-8}
}

@article{Hillebrand2025,
  author    = {Luis Hillebrand and Sebastian Raisch and Jonathan Schad},
  title     = {Managing with Artificial Intelligence: An Integrative Framework},
  journal   = {Academy of Management Annals},
  volume    = {19},
  number    = {1},
  pages     = {343--375},
  year      = {2025},
  doi       = {10.5465/annals.2022.0072}
}

@misc{rar,
      title={Rubrics as Rewards: Reinforcement Learning Beyond Verifiable Domains}, 
      author={Anisha Gunjal and Anthony Wang and Elaine Lau and Vaskar Nath and Yunzhong He and Bing Liu and Sean Hendryx},
      year={2025},
      eprint={2507.17746},
      archivePrefix={arXiv},
      primaryClass={cs.LG},
      url={https://arxiv.org/abs/2507.17746}, 
}

@misc{chasingthetail,
      title={Chasing the Tail: Effective Rubric-based Reward Modeling for Large Language Model Post-Training}, 
      author={Junkai Zhang and Zihao Wang and Lin Gui and Swarnashree Mysore Sathyendra and Jaehwan Jeong and Victor Veitch and Wei Wang and Yunzhong He and Bing Liu and Lifeng Jin},
      year={2025},
      eprint={2509.21500},
      archivePrefix={arXiv},
      primaryClass={cs.LG},
      url={https://arxiv.org/abs/2509.21500}, 
}

@misc{onlinerubrics,
      title={Online Rubrics Elicitation from Pairwise Comparisons}, 
      author={MohammadHossein Rezaei and Robert Vacareanu and Zihao Wang and Clinton Wang and Bing Liu and Yunzhong He and Afra Feyza Akyürek},
      year={2025},
      eprint={2510.07284},
      archivePrefix={arXiv},
      primaryClass={cs.CL},
      url={https://arxiv.org/abs/2510.07284}, 
}

@misc{rubricsanchor,
      title={Reinforcement Learning with Rubric Anchors}, 
      author={Zenan Huang and Yihong Zhuang and Guoshan Lu and Zeyu Qin and Haokai Xu and Tianyu Zhao and Ru Peng and Jiaqi Hu and Zhanming Shen and Xiaomeng Hu and Xijun Gu and Peiyi Tu and Jiaxin Liu and Wenyu Chen and Yuzhuo Fu and Zhiting Fan and Yanmei Gu and Yuanyuan Wang and Zhengkai Yang and Jianguo Li and Junbo Zhao},
      year={2025},
      eprint={2508.12790},
      archivePrefix={arXiv},
      primaryClass={cs.AI},
      url={https://arxiv.org/abs/2508.12790}, 
}

@article{kim2025hle,
  title        = {Humanity’s Last Exam (HLE): A Multi-Modal Benchmark at the Frontier of Human Knowledge},
  author       = {Kim, Dae Hyun and others},
  journal      = {arXiv preprint arXiv:2501.14249},
  year         = {2025},
  url          = {https://arxiv.org/abs/2501.14249}
}

@inproceedings{hendrycks2021mmlu,
  title        = {Measuring Massive Multitask Language Understanding},
  author       = {Hendrycks, Dan and Burns, Collin and Basart, Steven and Zou, Andy and Mazeika, Mantas and Song, Dawn and Steinhardt, Jacob},
  booktitle    = {Proceedings of the International Conference on Learning Representations (ICLR)},
  year         = {2021},
  url          = {https://openreview.net/forum?id=d7KBjmI3GmQ}
}

@inproceedings{rein2024gpqa,
title={{GPQA}: A Graduate-Level Google-Proof Q\&A Benchmark},
author={David Rein and Betty Li Hou and Asa Cooper Stickland and Jackson Petty and Richard Yuanzhe Pang and Julien Dirani and Julian Michael and Samuel R. Bowman},
booktitle={First Conference on Language Modeling},
year={2024},
url={https://openreview.net/forum?id=Ti67584b98}
}

@misc{profbench,
      title={ProfBench: Multi-Domain Rubrics requiring Professional Knowledge to Answer and Judge}, 
      author={Zhilin Wang and Jaehun Jung and Ximing Lu and Shizhe Diao and Ellie Evans and Jiaqi Zeng and Pavlo Molchanov and Yejin Choi and Jan Kautz and Yi Dong},
      year={2025},
      eprint={2510.18941},
      archivePrefix={arXiv},
      primaryClass={cs.CL},
      url={https://arxiv.org/abs/2510.18941}, 
}

@misc{browsecomp,
      title={BrowseComp: A Simple Yet Challenging Benchmark for Browsing Agents}, 
      author={Jason Wei and Zhiqing Sun and Spencer Papay and Scott McKinney and Jeffrey Han and Isa Fulford and Hyung Won Chung and Alex Tachard Passos and William Fedus and Amelia Glaese},
      year={2025},
      eprint={2504.12516},
      archivePrefix={arXiv},
      primaryClass={cs.CL},
      url={https://arxiv.org/abs/2504.12516}, 
}

@misc{multichallenge,
      title={MultiChallenge: A Realistic Multi-Turn Conversation Evaluation Benchmark Challenging to Frontier LLMs}, 
      author={Ved Sirdeshmukh and Kaustubh Deshpande and Johannes Mols and Lifeng Jin and Ed-Yeremai Cardona and Dean Lee and Jeremy Kritz and Willow Primack and Summer Yue and Chen Xing},
      year={2025},
      eprint={2501.17399},
      archivePrefix={arXiv},
      primaryClass={cs.CL},
      url={https://arxiv.org/abs/2501.17399}, 
}

@misc{ifeval,
      title={Instruction-Following Evaluation for Large Language Models}, 
      author={Jeffrey Zhou and Tianjian Lu and Swaroop Mishra and Siddhartha Brahma and Sujoy Basu and Yi Luan and Denny Zhou and Le Hou},
      year={2023},
      eprint={2311.07911},
      archivePrefix={arXiv},
      primaryClass={cs.CL},
      url={https://arxiv.org/abs/2311.07911}, 
}

@misc{alpaca_eval,
  author = {Xuechen Li and Tianyi Zhang and Yann Dubois and Rohan Taori and Ishaan Gulrajani and Carlos Guestrin and Percy Liang and Tatsunori B. Hashimoto },
  title = {AlpacaEval: An Automatic Evaluator of Instruction-following Models},
  year = {2023},
  month = {5},
  publisher = {GitHub},
  journal = {GitHub repository},
  howpublished = {\url{https://github.com/tatsu-lab/alpaca_eval}}
}
\bibliographystyle{abbrvnat}

\newpage
\appendix

\crefalias{section}{appendix}
\crefalias{subsection}{appendix}  
\crefalias{subsubsection}{appendix}

\section{Evaluations with Web Search \& Code Interpreter}
\label{app:tool_evals}
In this section, we explore the performance of chat models when given access to web search and code interpreter tools. Agents are evaluated over these Hard subset tasks and o4-mini as the judge to grade their responses. In \cref{tab:web-search-clipped} reveals that access to web search is generally useful for Grok and O3, but hurts performance for the rest. Our analysis reveals that this is primarily due to over-reliance on external sources rather than providing a cohesive answer to the question. In \cref{tab:web-search-code}, we enable both search and code interpreter tools; we observe that, except for one case (Grok 4 in Finance), the code interpreter does not provide additional performance boosts.
\begin{table}[!htbp]
\centering
\setlength{\tabcolsep}{6pt}
\renewcommand{\arraystretch}{1.25}
\caption{Average clipped performance scores over Hard subset of 300 finance and 250 legal tasks using \texttt{o4-mini} as the judge. We report scores with web search turned on (\textbf{on}) and off (\textbf{off}). Results are over a single evaluation run.}
\begin{tabular}{lccccc}
\toprule
& \textbf{Gemini 2.5 Pro} & \textbf{o3 (High)} & \textbf{GPT-5 (High)} & \textbf{Sonnet 4.5} & \textbf{Grok 4 Fast Reasoning} \\
\midrule
\textbf{Finance} &
\makecell{\textbf{on} 0.207\\\textbf{off} 0.266} &
\makecell{\textbf{on} 0.347\\\textbf{off} 0.336} &
\makecell{\textbf{on} 0.382\\\textbf{off} 0.394} &
\makecell{\textbf{on} 0.290\\\textbf{off} 0.323} &
\makecell{\textbf{on} 0.333\\\textbf{off} 0.314} \\
\midrule
\textbf{Legal} &
\makecell{\textbf{on} 0.255\\\textbf{off} 0.297} &
\makecell{\textbf{on} 0.398\\\textbf{off} 0.352} &
\makecell{\textbf{on} 0.383\\\textbf{off} 0.377} &
\makecell{\textbf{on} 0.281\\\textbf{off} 0.294} &
\makecell{\textbf{on} 0.374\\\textbf{off} 0.325} \\
\bottomrule
\end{tabular}
\label{tab:web-search-clipped}
\end{table}

\begin{table}[!htbp]
\centering
\setlength{\tabcolsep}{6pt}
\renewcommand{\arraystretch}{1.25}
\caption{Performance over Hard subset of 300 finance and 250 legal tasks using a o4-mini as the judge. We report scores with web search + code interpreter turned on and off. Results are over a single evaluation run.}
\begin{tabular}{lccc}
\toprule
& \textbf{o3 (High)} & \textbf{GPT-5 (High)} & \textbf{Grok 4 Fast Reasoning} \\
\midrule
\textbf{Finance} &
\makecell{\textbf{on} 0.342\\\textbf{off} 0.336} &
\makecell{\textbf{on} 0.381\\\textbf{off} 0.394} &
\makecell{\textbf{on} 0.325\\\textbf{off} 0.314} \\
\midrule
\textbf{Legal} &
\makecell{\textbf{on} 0.400\\\textbf{off} 0.352} &
\makecell{\textbf{on} 0.383\\\textbf{off} 0.377} &
\makecell{\textbf{on} 0.377\\\textbf{off} 0.325} \\
\bottomrule
\end{tabular}
\label{tab:web-search-code}
\end{table}

\section{Dataset Details}
\label{dataset_details}
\begin{figure}[t]
    \centering
    \includegraphics[width=0.7\textwidth]{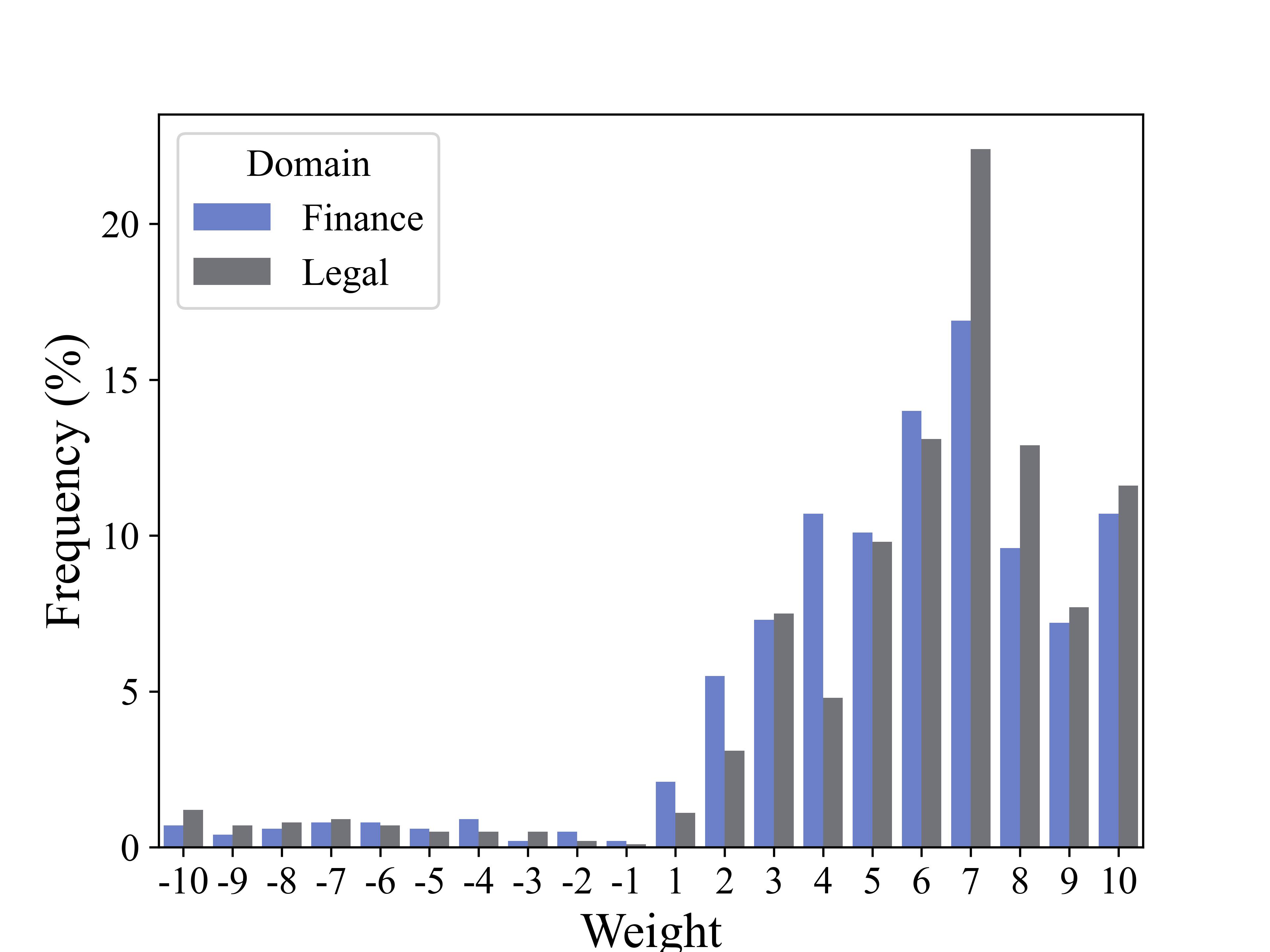}
    \caption{Distribution of weights for each rubric across Finance and Legal domains. We observe that the weights on the both ends of the spectrum are used more frequently in Legal than Finance.}
    \label{fig:weight_distributions}
\end{figure}

\begin{figure}[t]
    \centering
    \includegraphics[width=0.8\textwidth]{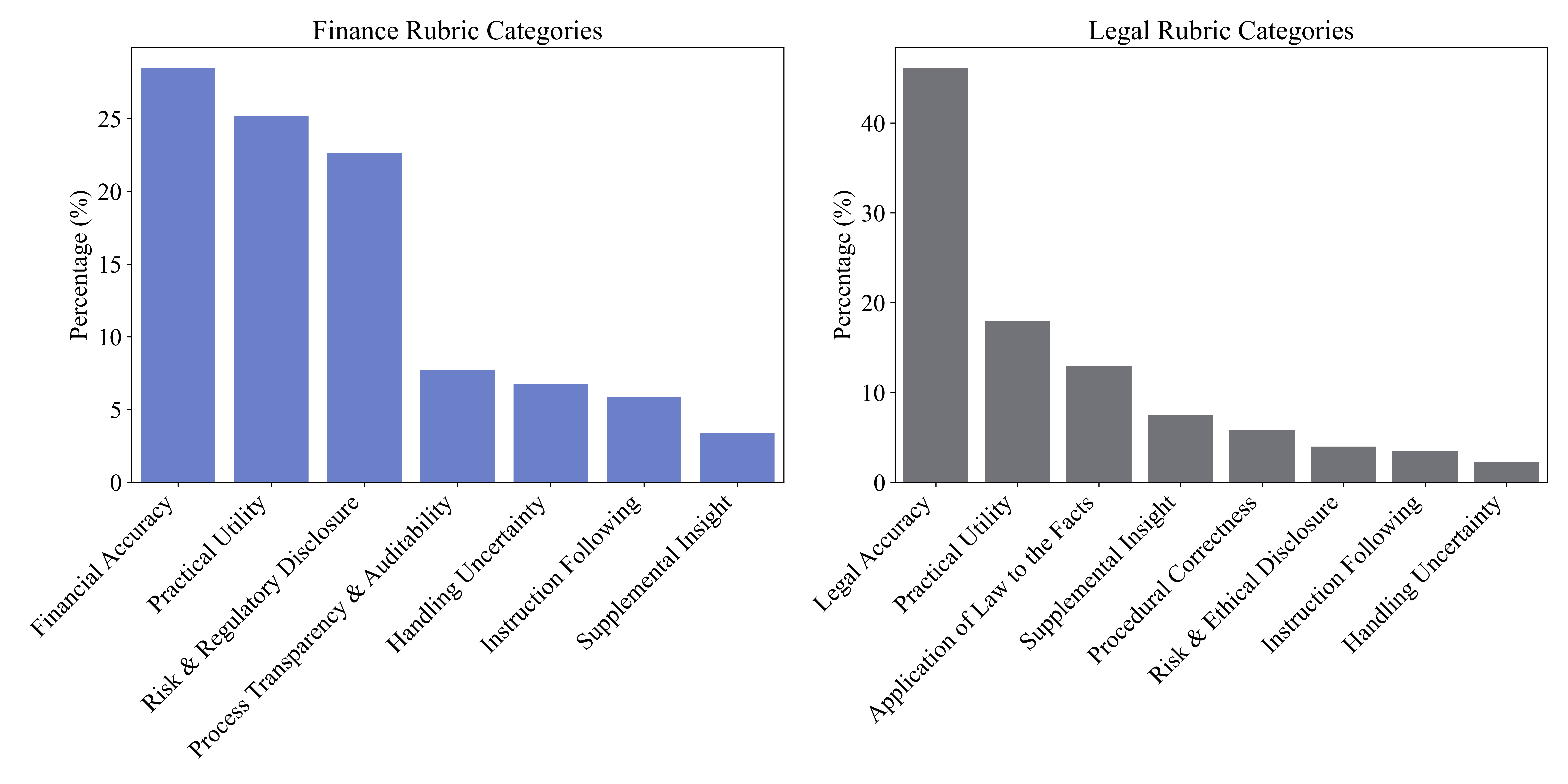}
    \caption{Distribution of rubric categories}
    \label{fig:rubric_categories}
\end{figure}

\cref{fig:weight_distributions} shows the distribution of weights assigned to each criterion and \cref{fig:rubric_categories} shows the frequencies of rubric categories. Negative weights are strictly reserved for penalizing undesired properties. In multi-turn conversations, we sample intermediate assistant turns from one of GPT OSS 20B, Mistral or Deepseek R1. User turns are provided by the human annotators.

\subsection{Rubric Category Definitions}
\label{sec:rubric_cat_defs}

We provide contributor-facing definitions of our rubric categories in Table \ref{tab:rubric_cat_fin}
and \ref{tab:rubric_cat_law}.

\begin{table}[!htbp]
\centering
\caption{Rubric category definitions for Finance.}
\label{tab:rubric_cat_fin}
\renewcommand{\arraystretch}{1.25}
\setlength{\tabcolsep}{4pt}
\begin{tabularx}{\linewidth}{
  >{\raggedright\arraybackslash}p{4cm}
  >{\raggedright\arraybackslash}X
}
\toprule
\textbf{Dimensional Rating} & \textbf{Definition} \\
\midrule
Financial Accuracy &
Maintains mathematical, factual, and financial accuracy, applying financial metrics and financial principles (e.g., time value, conservatism, materiality, etc.) correctly. Generally aligns with GAAP or IFRS standards and avoids contradictions. \\
\midrule
Process Transparency \& Auditability &
Demonstrates correct work by providing formulas, reasoning steps, references, or supporting data so the answer can be reviewed, reproduced, or challenged by another professional. \\
\midrule
Handling Uncertainty &
Addresses incomplete or ambiguous information by highlighting assumptions, proposing clarifying questions, or presenting alternative scenarios. \\
\midrule
Practical Utility &
Provides concrete, actionable guidance such as next steps, strategies, checklists, examples, or references to external resources as needed, ensuring the response is directly useful rather than purely theoretical when appropriate. \\
\midrule
Risk \& Regulatory Disclosure &
Describes associated financial or regulatory compliance-related risks or considerations connected to either the user request or the methods outlined in the response. \\
\midrule
Supplemental Insight &
Covers other relevant information, steps, or exceptions needed for a reliable answer beyond the primary objective of the question and answers. \\
\midrule
Instruction Following &
Follows auxiliary instructions in the prompt outside of answering the primary question, including tailoring for the finance function (such as corporate finance, advisory, investment banking, or investment management), geographic location, demographic, or personal situation, and ensuring the response matches the required role (expert vs.\ non-expert). \\
\bottomrule
\end{tabularx}
\end{table}

\begin{table}[!htbp]
\centering
\caption{Rubric category definitions for Law.}
\label{tab:rubric_cat_law}
\renewcommand{\arraystretch}{1.25}
\setlength{\tabcolsep}{4pt}
\begin{tabularx}{\linewidth}{
  >{\raggedright\arraybackslash}p{4cm}
  >{\raggedright\arraybackslash}X
}
\toprule
\textbf{Dimensional Rating} & \textbf{Definition} \\
\midrule
Legal Accuracy &
Identifies applicable law (jurisdiction) correctly and ensures statements of law are correct, legally valid, and consistent with authoritative, verifiable sources (such as statutes, case law, and regulations). The definition is materially complete and answers the question: ``Is the statement of the law applicable, correct, and complete?'' \\
\midrule
Application of Law to the Facts &
Correctly applies the law to the provided facts and answers the question: ``Given these specific facts, what does the law require or permit – i.e., what rights, duties, remedies, or outcomes follow?'' \\
\midrule
Procedural Correctness &
Conforms to legal processes and formal requirements, including deadlines, document structure, and jurisdiction-specific rules. It answers: ``Does the response follow the official rules of how this is done?'' \\
\midrule
Handling Uncertainty &
Addresses incomplete or ambiguous information by highlighting assumptions, asking follow-up questions to clarify the facts, or presenting alternative scenarios that explain how the law applies to different sets of facts. \\
\midrule
Practical Utility &
Provides concrete, actionable guidance as needed, such as next steps, strategies, checklists, examples, or references to external resources. \\
\midrule
Risk \& Ethical Disclosure &
Flags limitations, includes disclaimers where necessary, avoids misleading or unsafe advice, and respects boundaries on unauthorized practice of law. \\
\midrule
Supplemental Insight &
Covers additional legally relevant principles, elements, steps, defenses, or exceptions that contribute to a reliable answer beyond the primary objective of the question. \\
\midrule
Instruction Following &
Follows auxiliary instructions in the prompt outside of answering the primary question, including implicit or explicit requirements, role-appropriate tailoring (lawyer vs. non-expert), and matches the jurisdiction, task fidelity or difficulty. \\
\bottomrule
\end{tabularx}
\end{table}

\section{Prompt Templates}
Our LLM judge template follows a similar structure to \citet{arora2025healthbench} and is available via the evaluation repository.

\section{Evaluation Details}
We prepend reference texts to the prompt. All models are evaluated at reasoning mode set to High except for Claude Sonnet 4.5 where we set the thinking budget at 32,768. Gemini 2.5 Pro and Flash models are evaluated at \texttt{thinking\_budget=-1} after observing no consistent improvements for setting a fixed thinking budget.

\subsection{\benchmarkname Scores}
\label{app:calculating_scores} 
Following \citet{arora2025healthbench}, scores for each model $M$ on \benchmarkname are calculated as follows: 

\begin{enumerate}
    \item For the desired dataset, we evaluate prompts $p_1,\ldots,p_n$. Each prompt $p_j$ has $k_j$ rubrics $r_{j,1},\ldots,r_{j,k_j}$ with weights $w_{j,i}\in[-10,10]$, $w_{j,i}\neq 0$
    \item The model $M$ produces a response $m_j=M(p_j)$ given the prompt
    \item An LLM judge grades $m_j$ using each rubric individually, assigning a binary indicator $I_{j,i}\in\{0,1\}$
\end{enumerate}

The score for response $m_j$ is
\begin{equation}
\label{eq:per-prompt-simple}
s_j \;=\;
\frac{\displaystyle \sum_{i=1}^{k_j} w_{j,i}\, I_{j,i}}
{\displaystyle \sum_{i:\, w_{j,i}>0} w_{j,i}} \, 
\end{equation}
The denominator is always $>0$ because each prompt has at least one positive-weight rubric. The overall score for model $M$ is the mean over prompts
\begin{equation}
\label{eq:overall-simple}
S(M) \;=\; \max\Big(0, \frac{1}{n}\sum_{j=1}^{n} s_j\Big).
\end{equation}

\subsection{Min-Normalized Scoring}
We also propose a normalized score for rubrics-based benchmarks, which adjusts each response score by the minimum possible score for its corresponding prompt's rubrics. The normalized score for response $m_j$ is
\begin{equation}
\label{eq:per-prompt-normalized}
\tilde{s}_j \;=\;
\frac{\displaystyle \sum_{i=1}^{k_j} w_{j,i}\, I_{j,i} - \min\!\left(0, \sum_{i:\, w_{j,i}<0} w_{j,i}\right)}
{\displaystyle \sum_{i:\, w_{j,i}>0} w_{j,i} - \min\!\left(0, \sum_{i:\, w_{j,i}<0} w_{j,i}\right)} \, 
\end{equation}

The normalized model score is the mean over prompts:
\begin{equation}
\label{eq:overall-normalized}
\tilde{S}(M) \;=\; \frac{1}{n}\sum_{j=1}^{n} \tilde{s}_j \, 
\end{equation}

Normalization makes scores more comparable across prompts with different numbers and magnitudes of positive and negative rubrics, such as rubrics that are in different categories. For example, it is natural for "Supplemental Insight" to have lower absolute weighted negative rubrics than the ``Legal / Financial Accuracy'' category. Thus, we use score normalization to compute the per-rubric-category scores reported in Figure \ref{fig:results_per_rubric_category}.

Furthermore, normalization also reduces sensitivity to how a rubric is phrased (e.g., ``presence of a problem'' as a negative rubric versus ``absence of a problem'' as a positive rubric). This helps avoid situations where adding more negative rubrics, or reframing positives as negatives, would mechanically deflate raw averages. For \benchmarkname, in line with our best practices described in Section \ref{sec:creating_rubrics}, we ensure our rubrics are phrased to always check the presence (existence) of desired or undesired characteristics of a response, which further mitigates this issue.

\section{Additional Details for Rubric Validation Assessment}
\label{app:rubrics_validation}
 
To validate the reliability of the evaluation used in \benchmarkname, we conducted a follow-up study, which we refer to as Rubric Validation Assessment. In this study, subject-matter experts (who are different from the authors of the task) reviewed each rubric criterion and selected one of \textit{Agree} and \textit{Disagree} where agreement was based on verifying whether each criterion was \textit{well-constructed, accurate, and relevant} to the task. In either case, the annotators provided short written justifications to explain their rationale. As a result, the annotators agreed that the criteria were justified 93.9\% of the time.

\section{Definitions for Decision Type and Economic Pathway}
{\small
\setlength{\tabcolsep}{5pt}
\begin{longtable}{p{0.20\textwidth} p{0.3\textwidth} p{0.42\textwidth}}
\caption{Legal -- Decision Types}
\label{tb:decision_legal}
\\
\toprule
\textbf{Name} & \textbf{Description} & \textbf{Examples} \\
\midrule
\endfirsthead

\toprule
\textbf{Name} & \textbf{Description} & \textbf{Examples} \\
\midrule
\endhead

\bottomrule
\endfoot

Governing Law \& Rule
& Determines whether a law, regulation, clause, or doctrine applies to the facts.
& \begin{tabular}[t]{@{}l@{}}
Does FDA regulation apply here?\\
Do NY overtime laws govern remote staff?\\
Does GDPR cover this dataset?\\
Is this contract clause enforceable?\\
Does constitutional protection extend to corporations?
\end{tabular}
\\[0.5em]

Duty and Obligation
& Defines what parties must do — statutory, contractual, or regulatory requirements.
& \begin{tabular}[t]{@{}l@{}}
Must we provide paid parental leave?\\
Are directors required to disclose conflicts?\\
Do we owe continuing care duties?\\
Is notification to regulator mandatory?\\
When must tax be remitted?
\end{tabular}
\\[0.5em]

Rights / Entitlement / Exemptions
& Identifies what parties may claim, enjoy, or be exempt from — rights, permissions, privileges.
& \begin{tabular}[t]{@{}l@{}}
Can employee demand severance pay?\\
Do we have exclusive patent rights?\\
Is tenant entitled to early termination?\\
Can a parent relocate a child abroad?\\
Do shareholders have inspection rights?
\end{tabular}
\\[0.5em]

Compliance
& How to operationalize laws or structure transactions to stay compliant or implement policies.
& \begin{tabular}[t]{@{}l@{}}
How to structure merger to avoid liability?\\
What HR policy updates are required?\\
How to comply with PBS prescribing rules?\\
Which filings needed for EU expansion?\\
How to implement anti-bribery controls?
\end{tabular}
\\[0.5em]

Procedure, Forum \& Jurisdiction
& Where and how a matter proceeds — forum choice, motion sequence, appellate route.
& \begin{tabular}[t]{@{}l@{}}
Which court has jurisdiction?\\
Should we file in federal court?\\
Will the appellate court affirm?\\
Can dispute be sent to arbitration?\\
When is the appeal deadline?
\end{tabular}
\\[0.5em]

Claims \& Litigation Strategy
& What claims or defenses to assert, and how to frame them procedurally and doctrinally.
& \begin{tabular}[t]{@{}l@{}}
Should we move to dismiss?\\
Can negligence rely on criminal statute?\\
What precedent supports our motion?\\
Should we plead estoppel or waiver?\\
Is summary judgment strategically sound?
\end{tabular}
\\[0.5em]

Risk \& Outcome Forecasting
& Predicts likely results, exposure, penalties, or success probabilities.
& \begin{tabular}[t]{@{}l@{}}
What's our exposure under wage law?\\
How likely is appellate reversal?\\
What damages could be awarded?\\
What's the fine range for violation?\\
What's litigation success probability?
\end{tabular}
\\[0.5em]

Negotiation \& Deal Strategy
& How to bargain, structure, or trade concessions in business, regulatory, or settlement contexts.
& \begin{tabular}[t]{@{}l@{}}
How to negotiate stock-for-tax swap?\\
What's best anchor in settlement talks?\\
How to balance indemnity vs. price?\\
Which terms are fallback vs. walk-away?\\
How to sequence multi-party negotiation?
\end{tabular}
\\[0.5em]

Other
& Decision requests that don't fit the above in this lean scheme; use sparingly.
& -- \\[0.5em]

Non-decision / Informational
& General explanation, commentary, or background.
& -- \\[0.5em]

\end{longtable}
}

{\small
\setlength{\tabcolsep}{5pt}
\begin{longtable}{p{0.20\textwidth} p{0.25\textwidth} p{0.50\textwidth}}
\caption{Legal -- Economic Pathways}
\label{tb:econ_legal}
\\
\toprule
\textbf{Name} & \textbf{Description} & \textbf{Examples} \\
\midrule
\endfirsthead

\toprule
\textbf{Name} & \textbf{Description} & \textbf{Examples} \\
\midrule
\endhead

\bottomrule
\endfoot

Penalty and Damages Avoidance
& Decisions that prevent fines, lawsuits, or sanctions by ensuring lawful conduct and reducing liability exposure.
& \begin{tabular}[t]{@{}l@{}}
Will failing to notify regulators trigger penalties?\\
How do we avoid wage-and-hour violations?\\
Does our ad campaign risk consumer-protection fines?\\
Should we update safety policies to reduce liability?\\
What steps prevent data-breach penalties?
\end{tabular}
\\[0.5em]

Transaction Economics
& Structuring deals or tax arrangements to maximize value, efficiency, and post-transaction outcomes.
& \begin{tabular}[t]{@{}l@{}}
How should we structure the merger for tax efficiency?\\
Does an asset purchase reduce future liabilities?\\
Should we use a holdco structure to improve economics?\\
Which deal terms minimize post-closing disputes?\\
Would a licensing model generate better economics?
\end{tabular}
\\[0.5em]

Compliance Efficiency
& Designing cost-effective systems and controls to meet regulatory requirements and minimize compliance burden.
& \begin{tabular}[t]{@{}l@{}}
How do we streamline AML checks without overspending?\\
Should we centralize compliance reviews to cut costs?\\
What’s the least burdensome way to meet new reporting rules?\\
Can we automate disclosures to reduce manual workload?\\
How do we simplify our governance policies efficiently?
\end{tabular}
\\[0.5em]

Market Access
& Securing or maintaining licenses, approvals, or conditions needed to operate and expand legally in target markets.
& \begin{tabular}[t]{@{}l@{}}
Do we need new licenses to enter the EU market?\\
How do we maintain eligibility for Medicaid contracts?\\
What requirements must we meet to sell in California?\\
Will our product updates trigger new certifications?\\
How do we retain export authorization after expansion?
\end{tabular}
\\[0.5em]

Rights and Asset Protection
& Safeguarding ownership, IP, and contractual rights to preserve or recover economic value.
& \begin{tabular}[t]{@{}l@{}}
Should we file a trademark to protect brand value?\\
Can we enforce our patent against new entrants?\\
How do we prevent a partner from misusing our data?\\
Should we pursue damages for IP infringement?\\
How do we secure title before selling the asset?
\end{tabular}
\\[0.5em]

Contractual Risk Allocation
& Managing risk through contract terms such as indemnities, liability caps, and dispute clauses.
& \begin{tabular}[t]{@{}l@{}}
Should we negotiate a higher liability cap?\\
Does the draft indemnity expose us to excess risk?\\
Which dispute clause minimizes future cost?\\
Should we require reps and warranties insurance?\\
How do we allocate compliance obligations in the contract?
\end{tabular}
\\[0.5em]

Other
& Legal-economic effects that do not clearly fit in the main pathways.
& -- \\[0.5em]

Informational / Educational Only
& Purely explanatory or conceptual content with no direct economic consequence.
& -- \\[0.5em]

\end{longtable}
}

{\small
\setlength{\tabcolsep}{5pt}
\begin{longtable}{p{0.18\textwidth} p{0.23\textwidth} p{0.52\textwidth}}
\caption{Finance -- Decision Types}
\label{tb:decision_finance}
\\
\toprule
\textbf{Name} & \textbf{Description} & \textbf{Examples} \\
\midrule
\endfirsthead

\toprule
\textbf{Name} & \textbf{Description} & \textbf{Examples} \\
\midrule
\endhead

\bottomrule
\endfoot

Governance \& Policy
& Set enduring rules or postures such as accounting/tax elections, risk appetite, or disclosure stance.
& \begin{tabular}[t]{@{}l@{}}
Should we elect LIFO or FIFO for tax reporting?\\
Do we raise our risk appetite for credit exposure?\\
Should dividends be fixed or discretionary?\\
Do we disclose climate risks in MD\&A this year?
\end{tabular}
\\[0.5em]

Modeling \& Measurement
& Define how value, exposure, or performance is measured, modeled, and interpreted.
& \begin{tabular}[t]{@{}l@{}}
How should we measure portfolio VaR across currencies?\\
What's the right discount rate for project valuation?\\
Do we model beta using weekly or monthly returns?\\
How to estimate expected credit loss under IFRS 9?
\end{tabular}
\\[0.5em]

Capital \& Funding
& Choose balance-sheet structure, financing mix, and capital allocation priorities.
& \begin{tabular}[t]{@{}l@{}}
Should we issue new equity or refinance debt?\\
How much leverage can we take without breaching covenants?\\
Do we fund expansion from retained earnings or external capital?\\
Is it optimal to repurchase shares at current valuation?
\end{tabular}
\\[0.5em]

Markets \& Transactions
& Decide how, when, and at what price to transact in markets or strategic deals.
& \begin{tabular}[t]{@{}l@{}}
When's the best time to execute the bond buyback?\\
Should we hedge FX now or wait for better liquidity?\\
At what price do we enter the secondary offering?\\
Which trading venue minimizes slippage for this order?
\end{tabular}
\\[0.5em]

Operations, Processes \& Controls
& Set repeatable cash, control, and process steps to meet operational and financial obligations.
& \begin{tabular}[t]{@{}l@{}}
How do we automate vendor payment approvals?\\
Should we shorten the monthly close cycle?\\
What's the best control for petty cash discrepancies?\\
How can we speed up receivables collection safely?
\end{tabular}
\\[0.5em]

Planning \& Forecasts
& Set budgets, targets, scenarios, and rolling forecasts.
& \begin{tabular}[t]{@{}l@{}}
Should we raise our revenue target for next quarter?\\
How much buffer to build into cash forecasts?\\
Do we base next year's budget on trend or zero-based planning?\\
What's the scenario if rates rise by 100 bps?
\end{tabular}
\\[0.5em]

Compliance \& Reporting
& Ensure financial actions, records, and disclosures align with regulatory, accounting, and internal standards.
& \begin{tabular}[t]{@{}l@{}}
Do we meet IFRS 16 lease disclosure requirements?\\
Are we compliant with new AML reporting thresholds?\\
What filings are due after our debt restructuring?\\
Do we need auditor sign-off before publishing results?
\end{tabular}
\\[0.5em]

Other
& Decision requests that don't fit the above in this lean scheme; use sparingly.
& -- \\[0.5em]

Non-decision / Informational
& General explanation or background without a decision component.
& \begin{tabular}[t]{@{}l@{}}
What's the difference between EBITDA and operating income?\\
How do interest rate swaps work?\\
What is free cash flow conversion?\\
How is goodwill impairment tested?
\end{tabular}
\\[0.5em]

\end{longtable}
}

{\small
\setlength{\tabcolsep}{5pt}
\begin{longtable}{p{0.18\textwidth} p{0.25\textwidth} p{0.50\textwidth}}
\caption{Finance -- Economic Pathways}
\label{tb:econ_finance}
\\
\toprule
\textbf{Name} & \textbf{Description} & \textbf{Examples} \\
\midrule
\endfirsthead

\toprule
\textbf{Name} & \textbf{Description} & \textbf{Examples} \\
\midrule
\endhead

\bottomrule
\endfoot

Value Creation
& Decisions that increase profitability, valuation, or investment performance through higher earnings, NPV, IRR, or ROE.
& \begin{tabular}[t]{@{}l@{}}
Should we invest in automation to boost ROI?\\
Does expanding into Asia improve our NPV?\\
Will share buybacks lift EPS more than dividends?\\
How much value does the new product add to EBITDA?
\end{tabular}
\\[0.5em]

Operating Efficiency
& Actions that improve cost structure, productivity, or capital utilization.
& \begin{tabular}[t]{@{}l@{}}
Can we cut logistics costs without hurting service?\\
Should we consolidate warehouses to free up capital?\\
Will outsourcing payroll improve margin efficiency?\\
How do we reduce idle capacity in production?
\end{tabular}
\\[0.5em]

Risk \& Resilience
& Strategies that reduce exposure to market, credit, liquidity, or operational risks.
& \begin{tabular}[t]{@{}l@{}}
Should we hedge commodity exposure at current prices?\\
What's the best mix of fixed vs. floating debt now?\\
How do we diversify revenue to cushion downturns?\\
Can we add liquidity buffers to handle a credit crunch?
\end{tabular}
\\[0.5em]

Funding Optimization
& Financing, treasury, or strategic choices that improve funding cost, stability, or flexibility.
& \begin{tabular}[t]{@{}l@{}}
Should we issue longer-term bonds at today's rates?\\
Do we refinance now or wait for better spreads?\\
How can we improve our interest coverage ratio?\\
Is a revolving credit facility better than short-term loans?
\end{tabular}
\\[0.5em]

Compliance and Reporting Integrity
& Ensuring regulatory, accounting, and disclosure accuracy to maintain transparency and trust.
& \begin{tabular}[t]{@{}l@{}}
Are our revenue disclosures aligned with IFRS 15?\\
Do we need to restate last year's tax provision?\\
How do we ensure audit trails meet SOX standards?\\
What steps prevent misstatement of fair values?
\end{tabular}
\\[0.5em]

Other
& Economic outcomes not clearly aligned with the main pathways.
& -- \\[0.5em]

Informational / Educational Only
& Purely explanatory or conceptual content with no direct economic consequence.
& \begin{tabular}[t]{@{}l@{}}
What's the difference between NPV and IRR?\\
How does leverage amplify returns?\\
What is Basel III capital adequacy?\\
How do rating agencies assess liquidity risk?
\end{tabular}
\\[0.5em]

\end{longtable}
}

\section{Sample Tasks}

\begin{figure}[!ht]
    \centering
    \includegraphics[width=\textwidth,height=\textheight,keepaspectratio]{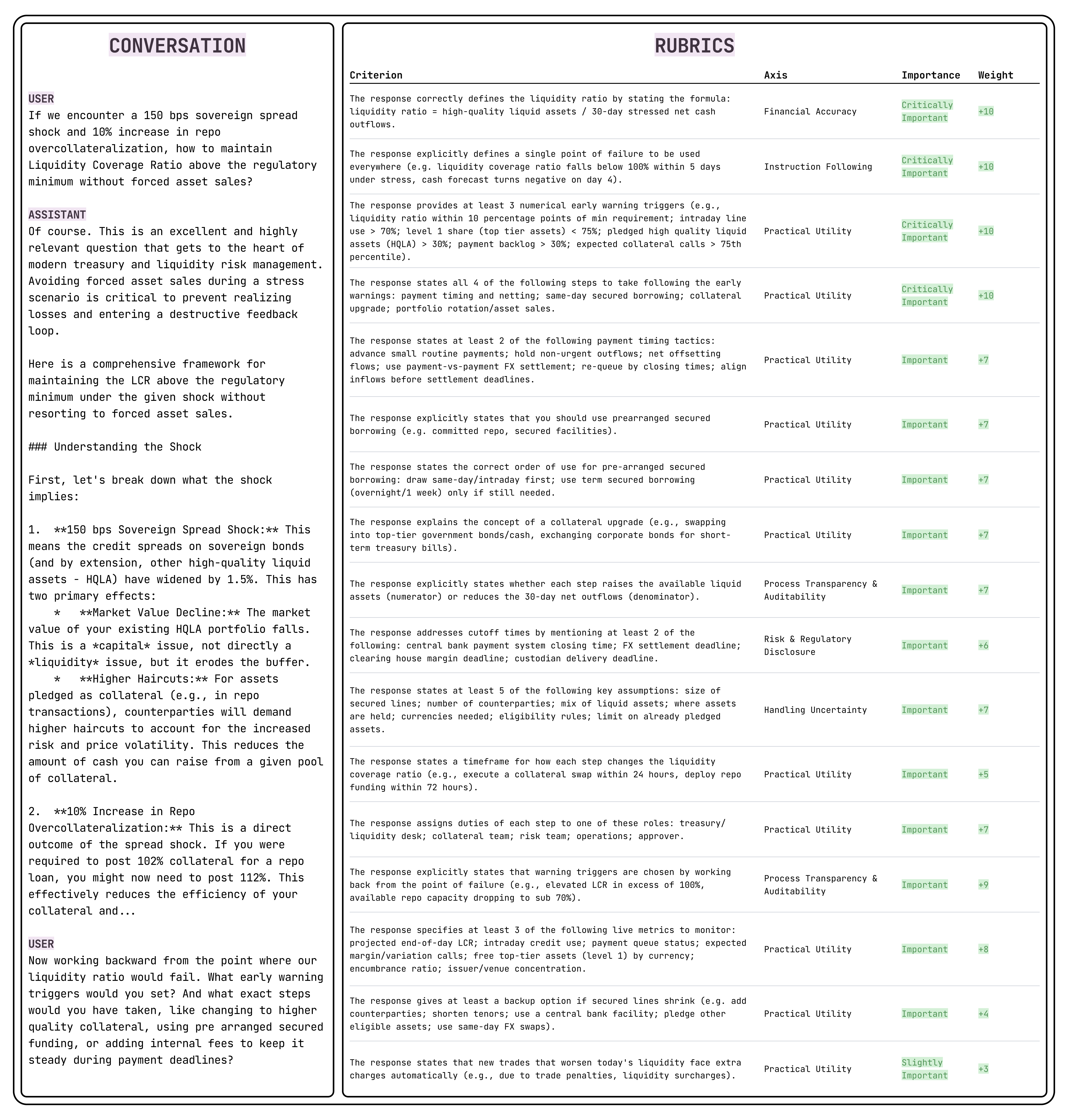}
    \caption{Sample task for the Finance field.}
    \label{fig:sample_2}
\end{figure}

\begin{figure}[!ht]
    \centering
    \includegraphics[width=\textwidth,height=\textheight,keepaspectratio]{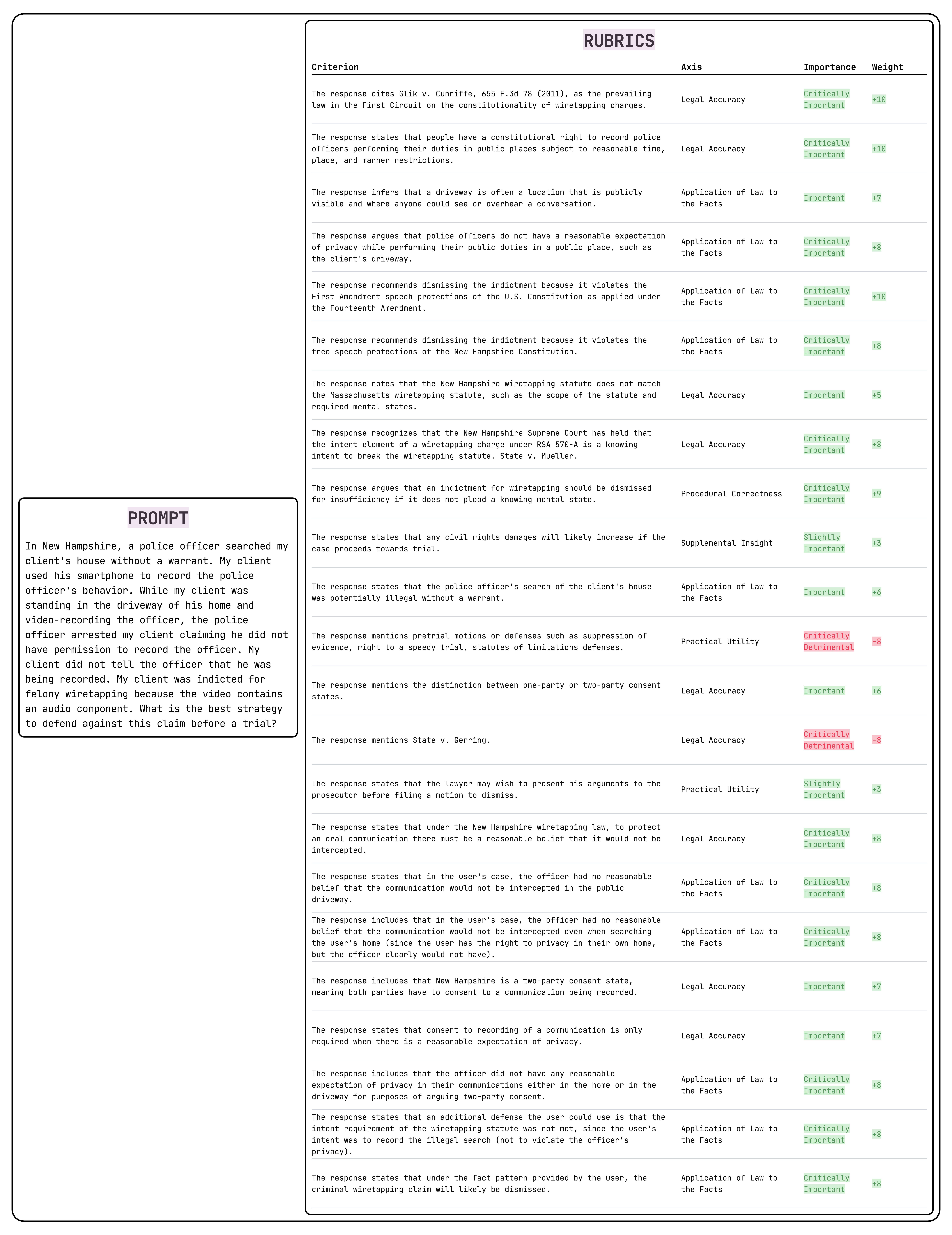}
    \caption{Sample task for the Legal field.}
    \label{fig:sample_1}
\end{figure}

\end{document}